\documentclass{article}

\usepackage{arxiv}

\usepackage[utf8]{inputenc}
\usepackage[T1]{fontenc}
\usepackage{url}
\usepackage{booktabs}
\usepackage{amsmath}
\usepackage{amsfonts}
\usepackage{nicefrac}
\usepackage{microtype}
\usepackage{xcolor}
\usepackage{multirow}
\usepackage{graphicx}
\usepackage{enumitem}
\usepackage{wrapfig}
\usepackage{natbib}
\usepackage{hyperref}

\title{Is One Layer Enough? Training a Single Transformer Layer Can Match Full-Parameter RL Training}

\author{%
  \makebox[0.31\textwidth][c]{%
    \begin{tabular}{c}
      \textbf{Zijian Zhang} \\
      \mdseries University of Minnesota \\
      \mdseries \texttt{zha00175@umn.edu}
    \end{tabular}%
  }%
  \makebox[0.31\textwidth][c]{%
    \begin{tabular}{c}
      \textbf{Rizhen Hu} \\
      \mdseries Peking University \\
      \mdseries \texttt{xshrz123@gmail.com}
    \end{tabular}%
  }%
  \makebox[0.31\textwidth][c]{%
    \begin{tabular}{c}
      \textbf{Athanasios Glentis} \\
      \mdseries University of Minnesota \\
      \mdseries \texttt{glent007@umn.edu}
    \end{tabular}%
  }%
  \vspace{1.5em}\\
  \makebox[0.31\textwidth][c]{%
    \begin{tabular}{c}
      \textbf{Dawei Li} \\
      University of Minnesota \\
      \texttt{li004678@umn.edu}
    \end{tabular}%
  }%
  \makebox[0.31\textwidth][c]{%
    \begin{tabular}{c}
      \textbf{Chung-Yiu Yau} \\
      University of Minnesota \\
      \texttt{cyau@umn.edu}
    \end{tabular}%
  }%
  \makebox[0.31\textwidth][c]{%
    \begin{tabular}{c}
      \textbf{Hongzhou Lin} \thanks{This work is independent of and outside of the work at Amazon} \\
      Amazon \\
      \texttt{hongzhou.lin89@gmail.com} 
    \end{tabular}%
  }%
  \vspace{1.5em}\\
  \makebox[0.31\textwidth][c]{%
    \begin{tabular}{c}
      \textbf{Mingyi Hong} \\
      University of Minnesota \\
      \texttt{mhong@umn.edu}
    \end{tabular}%
  }%
}

\renewcommand{\shorttitle}{Is One Layer Enough? Training a Single Transformer Layer Can Match Full-Parameter RL Training}
\renewcommand{\headeright}{Preprint}
\renewcommand{\undertitle}{Preprint}
\date{}

\begin{document}
\maketitle
\begin{abstract}
Reinforcement learning (RL) has become a central component of post-training large language models (LLMs), yet little is understood about how RL adaptation is distributed across transformer layers. Existing approaches typically update all model parameters uniformly, implicitly assuming that every layer contributes similarly to the gains obtained during RL post-training. In this work, we challenge this assumption through a systematic layer-wise study of RL training.

Surprisingly, we find that training {\bf a single transformer layer} can recover most of the gains achieved by full-parameter RL training, and in some cases even surpass it. To quantify this phenomenon, we introduce the quantity {\it layer contribution}, which measures the fraction of full RL improvement recovered by training a layer in isolation. Across seven models spanning two model families (Qwen3, Qwen2.5), three RL algorithms (GRPO, GiGPO, Dr.~GRPO), and multiple task domains including mathematical reasoning, code generation, and agentic decision-making, we observe a remarkably stable pattern: {\bf RL gains are highly concentrated in a small subset of, and in many cases even a single, transformer layers}. 
More strikingly, the same structural pattern consistently emerges: high-contribution layers concentrate in the middle of the transformer stack, while layers near the input and output ends contribute substantially less. The resulting layer rankings remain strongly correlated across datasets, tasks, model families, and RL algorithms.

Our findings have two important implications. First, they reveal a previously unrecognized structural property of RL post-training: most RL gains are concentrated in a small subset of transformer layers rather than being uniformly distributed throughout the network. Second, they suggest new opportunities for improving RL training. Guided by the above observation, we develop simple layer-aware training strategies that consistently outperform standard full-parameter RL training, while ensembles of layer-specialized models provide additional gains through complementary behaviors. Together, our results provide new insights into how RL modifies large language models and suggest a new perspective for understanding and improving RL post-training. 

\end{abstract}

\begin{figure}[t]
    \centering
    \includegraphics[width=\textwidth]{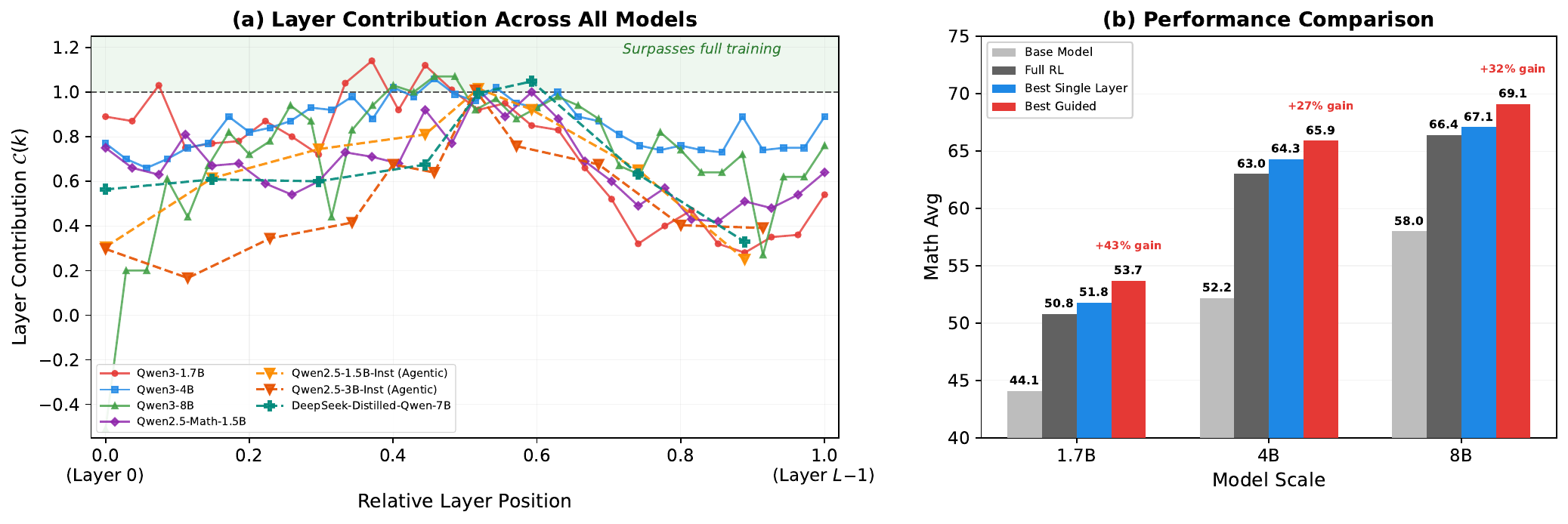}
    \caption{(a)~Layer contribution (defined in \S\ref{sec:layercontribution}) across all seven models studied in this work, plotted against depth-normalized relative layer position (0 = Layer~0, 1 = Layer~$L{-}1$). Each point corresponds to RL training performed on a single transformer layer while all remaining layers are frozen. The y-axis measures the fraction of full-parameter RL improvement recovered by training that layer alone. Solid lines denote models with full layer scans; dashed lines denote models where a representative subset of layers is trained. Despite spanning two model families, three RL algorithms, and two task domains (mathematical reasoning and agentic tasks), all seven models exhibit the same structure: layers around 40\%--60\% of network depth consistently achieve the highest contribution, with some surpassing full-parameter RL training (green-shaded region). (b)~Performance comparison on the three Qwen3 models (NuminaMath-CoT, math benchmarks). For each scale we show four bars: the base model, full-parameter RL, the best single-layer model, and \emph{Best Guided}---the best result among the layer contribution-guided strategies from \S\ref{sec:guiding} (either boosting the learning rate of high-contribution layers or selectively training only those layers). The percentage annotations indicate the additional gain of Best Guided beyond full-parameter RL, expressed relative to the total RL gain. Training only these high-contribution layers consistently outperforms full-parameter RL across all three scales. Note that in (b), the full-parameter RL and Best Guided bars are averaged over 3 evaluation runs (\S\ref{sec:guiding}); the full-parameter values therefore differ slightly from the single-run results reported in Table~\ref{tab:main}. }
    \label{fig:hero}
\end{figure}

\section{Introduction}

Reinforcement learning with verifiable rewards (RLVR) ~\citep{Guo_2025,yang2025qwen3technicalreport} has become a central component of post-training large language models (LLMs), driving substantial improvements in mathematical reasoning, code generation, and agentic decision-making ~\citep{yu2025dapoopensourcellmreinforcement, shao2024deepseekmathpushinglimitsmathematical}. While significant effort has been devoted to developing more effective RL objectives, reward models, and optimization algorithms, much less is known about a more fundamental question: {\bf where, within the network, do the gains from RL actually emerge?} Existing RL post-training methods update all transformer layers jointly, providing little visibility into how different parts of a pretrained model contribute to the improvements achieved through RL. Understanding this layer-wise structure is important not only for explaining how RL reshapes a pretrained language model, but also for revealing whether RL adaptation possesses an underlying structure that can be exploited to further improve RL post-training.

In this work, we conduct a systematic layer-wise investigation of RL post-training. Surprisingly, we find that a single transformer layer can often recover most of the gains achieved by full-parameter RL training, and can sometimes even outperform it. This finding challenges the intuition that RL improvements arise from coordinated adaptation across the entire network. Instead, our results suggest that much of the benefit of RL post-training is concentrated in a surprisingly small subset of transformer layers.

 Our study naturally connects to a growing body of work showing that pretrained LLMs are highly non-uniform across depth. Prior studies have demonstrated that transformer layers play markedly different roles: removing some layers causes severe degradation while others have relatively little effect~\citep{zhang2024investigatinglayerimportancelarge,song2026demystifyingrolesllmlayers,nepal2025layerimportancemathematicalreasoning}. Similar layer-wise heterogeneity has also been observed during supervised finetuning, where both parameter updates and adaptation behaviors vary substantially across layers~\citep{pan2024lisalayerwiseimportancesampling,shi2025understandinglayersignificancellm}. These observations have motivated a variety of layer-aware optimization strategies, including layer-wise sampling~\citep{pan2024lisalayerwiseimportancesampling}, adaptive layer selection~\citep{liu2026misamemoryefficientllmsoptimization,kumar2025adagradselectadaptivegradientguidedlayer}, and layer-aware alignment methods~\citep{shi2025understandinglayersignificancellm}. Notably, the layers identified as important often remain remarkably consistent across datasets and training settings~\citep{shi2025understandinglayersignificancellm}, suggesting that pretrained LLMs possess stable layer-wise structural organization. However, these studies have focused primarily on inference-time behavior and supervised finetuning. Whether RL post-training exhibits a similarly structured pattern has remained largely unexplored. %

To characterize this phenomenon, we conduct a systematic layer-wise study of RL post-training. For an LLM with $L$ transformer layers, we independently train each layer using RL while freezing all remaining layers, and compare the resulting improvement with that achieved by standard full-parameter RL training. We introduce a simple metric called {\it layer contribution}, which measures the fraction of full RL improvement that can be recovered by training a layer in isolation. This framework allows us to directly quantify the contribution of each layer to the gains achieved by RL post-training.

Our experiments reveal two striking findings. First, layer contributions vary dramatically across the network. The best individual layers recover up to 114\% of the gains achieved by full-parameter RL training, while the weakest layers recover less than 30\%.  Second, this variation is highly structured rather than random. Across seven models spanning two model families (Qwen3, Qwen2.5), three RL algorithms (GRPO, GiGPO, Dr.~GRPO), and three task domains including mathematical reasoning, code generation, and agentic decision-making, high-contribution layers consistently concentrate in the middle of the transformer stack, while layers near the input and output ends contribute substantially less. Further, for a fixed model, the per-layer contribution rankings themselves remain strongly correlated across training datasets (NuminaMath-CoT vs.\ DeepScaleR, Spearman $\rho=0.76$) and even across tasks (NuminaMath-CoT vs.\ DeepCoder, Spearman $\rho=0.59$; Figure~\ref{fig:cross_dataset}). %
Together, these findings point to a previously unrecognized structural property of RL post-training: most RL gains are concentrated in a small subset of transformer layers.

These findings have two important implications. First, they reveal a previously unrecognized structural property of RL post-training. Contrary to the intuition that RL improvements arise from coordinated adaptation across the entire network, our results suggest that much of the benefit of RL post-training is concentrated in a small and stable subset of transformer layers. Second, this structure can be exploited algorithmically. Guided by layer contribution, we develop simple layer-aware training strategies that prioritize high-contribution layers and consistently outperform standard full-parameter RL training. For example, on Qwen3-8B, training only the ten highest-contribution layers achieves 69.1\% average accuracy on mathematical reasoning benchmarks, compared to 66.4\% achieved by full-parameter RL training. Furthermore, models trained on different layers exhibit complementary problem-solving behaviors, and combining them through majority voting yields additional gains beyond the full-parameter baseline.

In summary, our main contributions are:
\begin{itemize}[leftmargin=*]
    \item \textbf{RL adaptation is concentrated.} We show that RL gains are highly unevenly distributed across transformer layers. Remarkably, training a single layer can recover most of the gains achieved by full-parameter RL training and can sometimes even surpass it. %
    \item \textbf{Layer contribution follows a consistent structure.} We introduce the notion of layer contribution and establish that high-contribution layers consistently concentrate in the middle of transformer networks across model scales, model families, RL algorithms, datasets, and task domains. %

    \item \textbf{Implications for RL post-training.} We show that the discovered layer structure can be exploited to improve RL post-training. Simple layer-aware strategies that prioritize high-contribution layers consistently outperform standard full-parameter training, while even a profiling-free heuristic that trains only middle layers achieves comparable or better performance. Furthermore, models trained on different layers exhibit complementary behaviors, and combining them through majority voting yields additional gains beyond the full-parameter baseline.
    
\end{itemize}

\section{Preliminaries}

\subsection{RLVR and GRPO}

Reinforcement Learning with Verifiable Rewards (RLVR) optimizes a
language model policy $\pi_\theta$ by maximizing expected reward on
tasks with objectively verifiable answers. Given a prompt $x$, the
model generates a response $y \sim \pi_\theta(\cdot|x)$, which is
evaluated by a reward function $r(x, y)$ that returns a binary signal
based on answer correctness.

In this work, we adopt Group Relative Policy Optimization
(GRPO)~\citep{shao2024deepseekmathpushinglimitsmathematical}, which estimates advantages without a
learned value network. For each prompt $x$, GRPO samples a group of
$G$ responses $\{y_1, \ldots, y_G\}$ from the current policy and
computes a group-normalized advantage for each response:
\begin{equation}
    \hat{A}_i = \frac{r(x, y_i) - \text{mean}(\{r(x, y_j)\}_{j=1}^G)}
    {\text{std}(\{r(x, y_j)\}_{j=1}^G)}
\end{equation}
The policy is then updated by maximizing a clipped surrogate
objective:
\begin{equation}
    \mathcal{L}_{\text{GRPO}}(\theta) = \mathbb{E}_{x, \{y_i\}}
    \left[ \frac{1}{G} \sum_{i=1}^{G} \left(\min \left(
    \rho_i \hat{A}_i, \;
    \text{clip}(\rho_i, 1-\epsilon, 1+\epsilon) \hat{A}_i
    \right) - \beta\, \mathbb{D}_{\mathrm{KL}}\!\left[\pi_\theta \,\|\, \pi_{\text{ref}}\right]\right) \right]
\end{equation}
where $\rho_i = \pi_\theta(y_i|x) / \pi_{\theta_{\text{old}}}(y_i|x)$
is the importance sampling ratio, $\beta$ is the KL penalty coefficient, and
$\pi_{\text{ref}}$ is the fixed reference policy (the initial model before RL).

\subsection{Single-Layer Training and Layer Contribution}
\label{sec:layercontribution}

To study how RL post-training interacts with different parts of a pretrained LLM, we adopt a controlled layer-wise training framework that isolates the adaptation behavior of each layer independently. For an LLM with $L$ Transformer layers $\{\theta_0, \theta_1, \ldots, \theta_{L-1}\}$, embedding parameters $\theta_{\text{emb}}$, and language model head $\theta_{\text{head}}$, standard full-parameter GRPO updates all parameters jointly by computing the gradient $\nabla_\theta \mathcal{L}_{\text{GRPO}}(\theta)$ over the full parameter set $\theta = \{\theta_{\text{emb}}, \theta_0, \ldots, \theta_{L-1}, \theta_{\text{head}}\}$. In our single-layer training framework, we instead isolate a single layer $\theta_k$ and update only its parameters:
\begin{equation}
    \theta_k \leftarrow \theta_k - \alpha \, \nabla_{\theta_k} \mathcal{L}_{\text{GRPO}}(\theta), \quad \text{all other parameters frozen.}
\end{equation}
Note that the gradient $\nabla_{\theta_k} \mathcal{L}_{\text{GRPO}}(\theta)$ is computed via backpropagation through the full network, so it depends on all layers; only the parameter \emph{update} is restricted to layer $k$. In practice, this is implemented in PyTorch by setting \texttt{requires\_grad=False} for all parameters except those in the target layer $\theta_k$. This procedure is repeated independently for every layer $k \in \{0, \ldots, L-1\}$, isolating each layer's ability to absorb RL-induced improvement. Each resulting model is then evaluated on the same set of in-domain benchmarks to obtain a performance score.

To quantify each layer's capacity to capture RL-induced improvement, we define \emph{layer contribution}. Let $S_k$ denote the in-domain performance of the model trained on layer $k$, measured as the average score across in-domain benchmarks. Let $S_{\text{base}}$ denote the performance of the original pretrained model without any RL training, and $S_{\text{full}}$ denote the performance of the model after standard full-parameter GRPO training. The layer contribution of layer $k$ is:
\begin{equation}\label{eq:layer}
    \mathcal{C}(k) = \frac{S_k - S_{\text{base}}}{S_{\text{full}} - S_{\text{base}}}.
\end{equation}
A layer contribution of 1.0 indicates that single-layer training fully matches the gain of full-parameter training; values above 1.0 indicate that it surpasses full-parameter training; values near 0 indicate that the layer fails to capture meaningful RL improvements.

\section{Measuring Layer Contribution in RLVR}
\label{sec:experiments}
We conduct systematic single-layer training experiments to examine whether different layers contribute equally to learning during RLVR. We study seven models spanning 1.5B to 8B parameters across two model families, three RL algorithms, and two task domains. We first describe the experimental setup and our protocol for ensuring fair comparison (\S\ref{sec:setup}), then present detailed layer contribution results on the Qwen3 models (\S\ref{sec:contribution}), and finally establish the consistency of these findings across datasets (\S\ref{sec:consistency}), model families, RL algorithms, and tasks (\S\ref{sec:generalization_family}).

\subsection{Experimental Setup and Fair Comparison}
\label{sec:setup}
{\paragraph{Models and training configurations.}
Our primary experiments use Qwen3-1.7B-Base (28 layers), Qwen3-4B-Base (36 layers), and Qwen3-8B-Base (36 layers)~\citep{yang2025qwen3technicalreport}, which have not undergone post-training and thus provide a clean starting point for isolating the effects of RL. For each of these three models, we perform GRPO training on every layer independently using NuminaMath-CoT~\citep{numina_math_datasets} as the training dataset, with all other layers, embedding parameters, and the language model head frozen. A full-parameter GRPO baseline trained under identical hyperparameters serves as the reference. We choose this setting for our primary experiments because its difficulty level is well-suited to Qwen3 base models.

Further, to validate the generality of our findings beyond the Qwen3 family, NuminaMath-CoT dataset and GRPO algorithm, we additionally conduct experiments on Qwen2.5-Math-1.5B (28 layers) trained with Dr.~GRPO~\citep{liu2025understanding}, and Qwen2.5-1.5B-Instruct (28 layers) and Qwen2.5-3B-Instruct (36 layers) trained with GiGPO~\citep{feng2025groupingrouppolicyoptimizationllm} on the agentic task ALFWorld~\citep{shridhar2021alfworldaligningtextembodied}, and DeepSeek-Distilled-Qwen-7B (28 layers) trained with GRPO on the Skywork~\citep{he2025skyworkopenreasoner1} mathematics dataset. 

In addition, to understand cross-dataset and cross-task consistency of layer contribution within a single and fixed model, beyond using the NuminaMath-CoT, we additionally train Qwen3-1.7B-Base with DeepScaleR~\citep{deepscaler2025} (mathematics) and DeepCoder~\citep{deepcoder2025} (coding).  

Table~\ref{tab:setup_summary} summarizes the seven models studied in this work. Note that in Table \ref{tab:setup_summary}, the last column "Layer Scan" indicates  if all the layers have been trained individually or only a selected subset of the layers are trained in our experiments. 

}

\begin{table}[t]
\centering
\caption{Summary of all models and training configurations studied in this work. }
\label{tab:setup_summary}
\vspace{2mm}
\resizebox{\textwidth}{!}{%
\begin{tabular}{llccllc}
\toprule
\textbf{Model} & \textbf{Family} & \textbf{Params} & \textbf{Layers} & \textbf{RL Algorithm} & \textbf{Task / Dataset} & \textbf{Layer Scan} \\
\midrule
Qwen3-1.7B-Base & Qwen3 & 1.7B & 28 & GRPO & Math / NuminaMath-CoT & Full \\
Qwen3-4B-Base & Qwen3 & 4B & 36 & GRPO & Math / NuminaMath-CoT & Full \\
Qwen3-8B-Base & Qwen3 & 8B & 36 & GRPO & Math / NuminaMath-CoT & Full \\
Qwen2.5-Math-1.5B & Qwen2.5 & 1.5B & 28 & Dr.~GRPO & Math / MATH & Full \\
Qwen2.5-1.5B-Instruct & Qwen2.5 & 1.5B & 28 & GiGPO & Agentic / ALFWorld & Partial \\
Qwen2.5-3B-Instruct & Qwen2.5& 3B & 36 & GiGPO & Agentic / ALFWorld & Partial \\
DeepSeek-Distilled-Qwen-7B & Qwen2.5 & 7B & 28 & GRPO & Math / Skywork & Partial \\
\bottomrule
\end{tabular}
}
\end{table}

\paragraph{Evaluation.}
For the primary Qwen3 experiments, we evaluate on 12 benchmarks spanning four categories: Math (MATH500, GSM8K, OlympiadBench, AMC) as the {\it in-domain evaluation}, and three out-of-domain categories, Code (HumanEval+, MBPP, LiveCodeBench), Reasoning (GPQA-Diamond, MMLU-Pro), and Language (C-Eval, IFEval, MGSM). The overall score is computed as the unweighted average of the four category scores. 

For the Qwen2.5-Math-1.5B experiment and DeepSeek-Distilled-Qwen-7B, we follow the evaluation protocol of~\citet{liu2025understanding} and report results on six math benchmarks (AIME~2024, AIME~2025, AMC, MATH500, Minerva~Math, OlympiadBench). For the agentic experiments, we evaluate on ALFWorld tasks. In the main text we report detailed in-domain results and out-of-distribution category averages; full per-benchmark breakdowns are provided in Appendix~\ref{app:full_results}.

{\paragraph{Fair comparison of training methods.}
A key effort in our study is to ensure that when comparing single-layer training and full-parameter training, any observed differences reflect genuine layer-level variation rather than artifacts of suboptimal hyperparameters or premature convergence. Our detailed protocol is given below. 

First, for each model, we tune the learning rate for the full-parameter baseline and select the value that yields the best performance; this ensures that the full-parameter reference is as strong as possible. Second, we apply this full-parameter-tuned learning rate to all single-layer training runs, so that no layer receives an unfair advantage or disadvantage from the learning rate choice. Third, all configurations, including full-parameter and single-layer, use identical hyperparameters for every other setting (batch size, KL coefficient, clip range, number of epochs) and are trained to convergence under the same training steps. This protocol ensures that when a single layer matches or surpasses full-parameter training, the comparison is rigorous: the full-parameter baseline is already at its best learning rate, and the single-layer run uses the same settings. Fourth, for a number of settings that have publicly available results using the same model, dataset, and methods, such as Dr.~GRPO and GiGPO, we also report the best publicly available results, so as to best anchor our own full-parameter experiments and  the performance achieved by layer-training.  %

Of course, since we only tuned the learning rate for full training but not the layer-wise training, a natural concern is whether low-contribution layers might improve with a larger learning rate, and if the high-contribution layer can be even better. We address this with a learning rate ablation study in Appendix~\ref{app:lr_ablation}, which shows that adjusting the learning rate does not change the layer contribution rankings. Full training details and hyperparameter tables are provided in Appendix~\ref{app:details}.

}

\subsection{Qwen3 Experiments: Layer Contribution Varies Dramatically}
\label{sec:contribution}

\begin{table}[t]
\centering
\caption{Per-layer training results on the three Qwen3 models. We report each in-domain math benchmark, three out-of-distribution category averages (Code, Reasoning, Language), and the overall average across all four categories. $\mathcal{C}_\text{math}$ and $\mathcal{C}_\text{all}$ denote layer contribution computed on the in-domain math average and the overall average, respectively. Complete per-layer results are in Appendix~\ref{app:full_results}.}
\label{tab:main}
\vspace{2mm}
\resizebox{\textwidth}{!}{%
\begin{tabular}{ll|ccccc|c|ccc|c|c}
\toprule
& & \multicolumn{6}{c|}{In-domain (Math)} & \multicolumn{3}{c|}{Out-of-distribution} & & \\
\cmidrule(lr){3-8} \cmidrule(lr){9-11}
\textbf{Model} & \textbf{Setting} & \textbf{MATH500} & \textbf{GSM8K} & \textbf{Olymp.} & \textbf{AMC} & \textbf{Avg} & $\mathcal{C}_\text{math}$ & \textbf{Code} & \textbf{Reas.} & \textbf{Lang.} & \textbf{Overall} & $\mathcal{C}_\text{all}$ \\
\midrule
\multirow{7}{*}{\shortstack{Qwen3-\\1.7B-Base}}
& Base     & 57.4 & 74.4 & 18.7 & 26.1 & 44.1 & 0.00 & 34.9 & 20.7 & 41.7 & 35.4 & 0.00 \\
& Full     & 64.0 & 82.0 & 26.9 & 30.2 & 50.8 & 1.00 & 33.5 & 22.6 & 48.2 & 38.8 & 1.00 \\
\cmidrule{2-13}
& Layer 10 & \textbf{68.6} & 80.5 & 27.3 & 30.8 & \textbf{51.8} & \textbf{1.14} & 34.6 & 21.9 & 47.2 & 38.9 & 1.03 \\
& Layer 12 & 65.6 & 81.3 & 27.3 & \textbf{32.4} & 51.6 & 1.12 & 36.2 & 21.5 & 47.4 & 39.2 & 1.12 \\
& Layer 1  & 64.4 & 79.4 & 25.9 & 30.2 & 50.0 & 0.87 & \textbf{40.0} & \textbf{22.7} & 47.1 & \textbf{39.9} & \textbf{1.32} \\
& Layer 7  & 64.0 & 80.1 & 24.9 & 29.0 & 49.5 & 0.80 & 38.1 & 22.4 & 46.5 & 39.1 & 1.09 \\
& Layer 24 & 60.6 & 74.8 & 21.2 & 27.6 & 46.1 & 0.28 & 30.6 & 21.6 & 44.2 & 35.6 & 0.06 \\
\midrule
\multirow{7}{*}{\shortstack{Qwen3-\\4B-Base}}
& Base     & 65.2 & 75.4 & 27.6 & 40.5 & 52.2 & 0.00 & 41.5 & 28.8 & 57.6 & 45.0 & 0.00 \\
& Full     & 77.2 & 91.9 & 38.4 & 47.1 & 63.7 & 1.00 & 48.8 & 32.4 & 62.9 & 51.9 & 1.00 \\
\cmidrule{2-13}
& Layer 16 & \textbf{79.4} & \textbf{92.0} & \textbf{40.3} & 45.5 & \textbf{64.3} & \textbf{1.06} & 51.9 & 33.0 & \textbf{64.4} & \textbf{53.4} & \textbf{1.22} \\
& Layer 14 & 78.4 & 90.3 & 39.9 & 46.5 & 63.8 & 1.02 & \textbf{52.9} & 31.6 & 63.3 & 52.9 & 1.14 \\
& Layer 11 & 76.6 & 90.5 & 36.2 & \textbf{47.6} & 62.7 & 0.92 & 51.3 & \textbf{33.8} & 62.8 & 52.7 & 1.12 \\
& Layer 24 & 77.0 & 89.5 & 38.5 & 43.6 & 62.2 & 0.87 & 47.4 & 31.9 & 61.7 & 50.8 & 0.84 \\
& Layer 2  & 73.8 & 87.7 & 35.3 & 42.4 & 59.8 & 0.66 & 49.1 & 31.9 & 62.3 & 50.8 & 0.84 \\
\midrule
\multirow{7}{*}{\shortstack{Qwen3-\\8B-Base}}
& Base     & 71.8 & 82.0 & 36.6 & 41.7 & 58.0 & 0.00 & 50.4 & 32.2 & 57.5 & 49.5 & 0.00 \\
& Full     & 80.0 & 92.3 & 42.8 & 50.8 & 66.5 & 1.00 & 53.7 & 35.5 & 63.7 & 54.9 & 1.00 \\
\cmidrule{2-13}
& Layer 16 & \textbf{80.4} & 91.8 & \textbf{44.1} & \textbf{52.0} & \textbf{67.1} & \textbf{1.07} & 54.5 & \textbf{35.5} & \textbf{68.8} & \textbf{56.5} & \textbf{1.30} \\
& Layer 15 & 79.8 & \textbf{92.8} & 40.6 & \textbf{52.7} & 66.5 & 1.00 & \textbf{56.8} & 34.0 & 68.9 & 56.5 & 1.30 \\
& Layer 8  & 77.0 & 89.0 & 41.5 & 50.9 & 64.6 & 0.78 & 54.3 & 34.4 & 62.0 & 53.8 & 0.80 \\
& Layer 2  & 72.8 & 84.6 & 38.5 & 43.0 & 59.7 & 0.20 & 51.4 & 33.7 & 59.4 & 51.1 & 0.30 \\
& Layer 0  & 61.8 & 79.6 & 31.0 & 42.4 & 53.7 & $-$0.51 & 44.0 & 34.0 & 63.0 & 48.7 & $-$0.15 \\
\bottomrule
\end{tabular}
}
\end{table}

\begin{figure}[t]
    \centering
    \includegraphics[width=\textwidth]{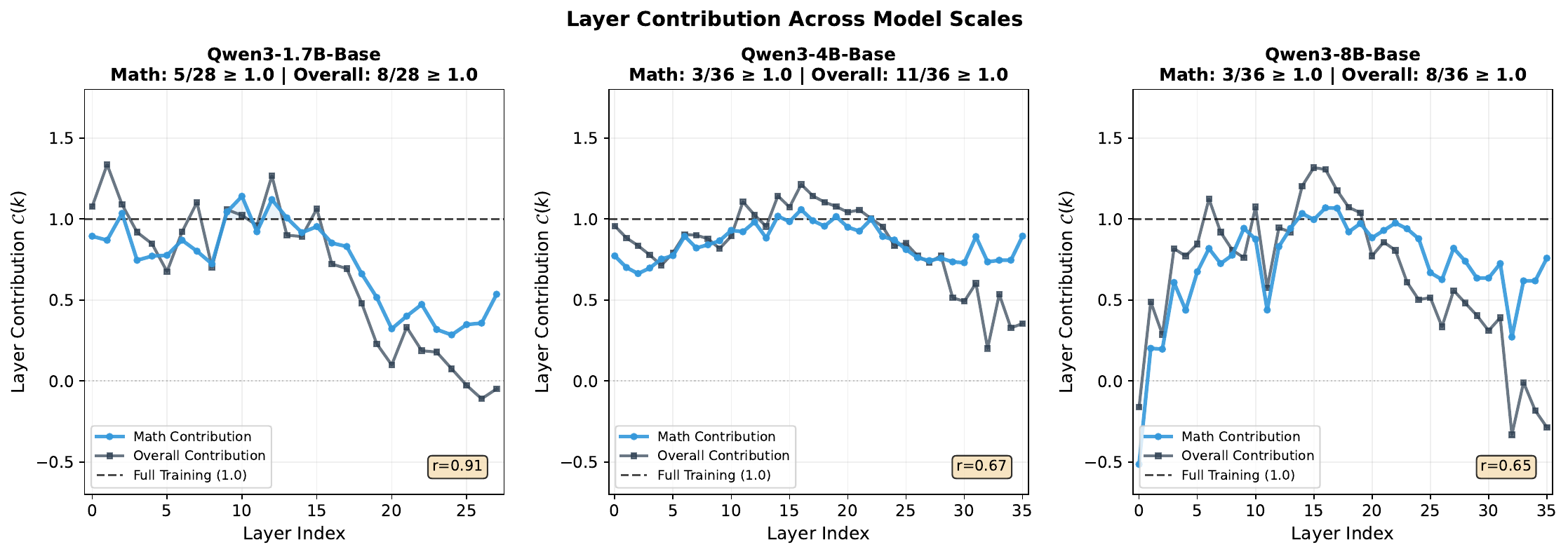}
    \caption{Layer contribution $\mathcal{C}(k)$ across model scales.
    Blue: math contribution (in-domain). Black: overall contribution
    (averaged across all capabilities). Dashed line indicates
    full-parameter training ($\mathcal{C}=1.0$). Each point represents
    one layer trained in isolation. Math and overall contribution
    closely track each other across layers (Pearson $r>0.6$ on 1.7B,4B and 8B), indicating that high-contribution layers
    achieve broad capability improvement rather than overfitting to
    the training objective. Across all three scales, middle layers
    consistently exhibit higher contribution.}
    \label{fig:contribution}
\end{figure}

We begin by presenting detailed layer contribution results on the three Qwen3 models, NuminaMath-CoT dataset and using GRPO. We will conduct independent training  on all 28 or 36 layers of these models. %
Figure~\ref{fig:contribution} presents the per-layer contribution across model scales. Details of the hyperparameter tuning, including the tuning of the full-parameter baseline, are provided in Appendix~\ref{app:details}.

On Qwen3-1.7B-Base, layer contribution ranges from 0.28 (Layer 24) to
1.14 (Layer 10), with 5 out of 28 layers exceeding 1.0 and 7 layers falling below 0.5. The fact that a single layer can capture the
entirety of the full training gain suggests that the effective change
induced by RLVR can be captured within the parameter subspace of a
single layer. Moreover, that some layers individually surpass
full-parameter training suggests that when all layers are trained
jointly, certain layers learn less effectively and may dilute the
overall improvement. At the other end, layers with contribution below
0.5 have limited capacity to learn from RL signals in isolation. The
variation across layers is not marginal but dramatic, with the best
layer capturing over four times the gain of the worst.

Similar patterns emerge on larger models. On Qwen3-4B-Base,
contribution ranges from 0.66 (Layer 2) to 1.06 (Layer 16), with 4 layers reaching or exceeding 1.0. On Qwen3-8B-Base, the best layers again reach
contributions above 1.0 (Layer 16, $\mathcal{C}=1.07$), while most
layers fall in the range of 0.6 to 1.0. A notable exception is Layer
0 on Qwen3-8B-Base, which exhibits a negative contribution
($\mathcal{C}=-0.51$), indicating that training this layer in
isolation actually degrades math performance below the base model.
Across all three model scales, middle layers consistently exhibit
higher contribution, while layers near the input and output ends
contribute less (Table~\ref{tab:main}).

Interestingly, high-contribution layers do not merely improve on the in-domain training objective, as they also improve out-of-distribution capabilities. To show this, we compute an \emph{overall} layer contribution $\mathcal{C}_\text{all}(k)$ using the same formula as Equation~\eqref{eq:layer} but replacing the in-domain math score with the overall score (the unweighted average of all four category scores: Math, Code, Reasoning, and Language; see Table~\ref{tab:main}). As shown in Figure~\ref{fig:contribution}, $\mathcal{C}_\text{all}$ closely tracks $\mathcal{C}_\text{math}$ across layers (Pearson $r>0.6$ on all three scales): layers that learn math effectively also tend to improve on out-of-distribution tasks including coding, reasoning, and language understanding. This indicates that single-layer training captures genuine, broad capability improvement rather than overfitting to the training objective, and that layer contribution reflects a general property of each layer rather than a task-specific one. 

\subsection{Qwen3 Experiments:    Layer Contribution is Consistent Across Datasets and Tasks}
\label{sec:consistency}
\begin{figure}[t]
    \centering
    \includegraphics[width=\textwidth]{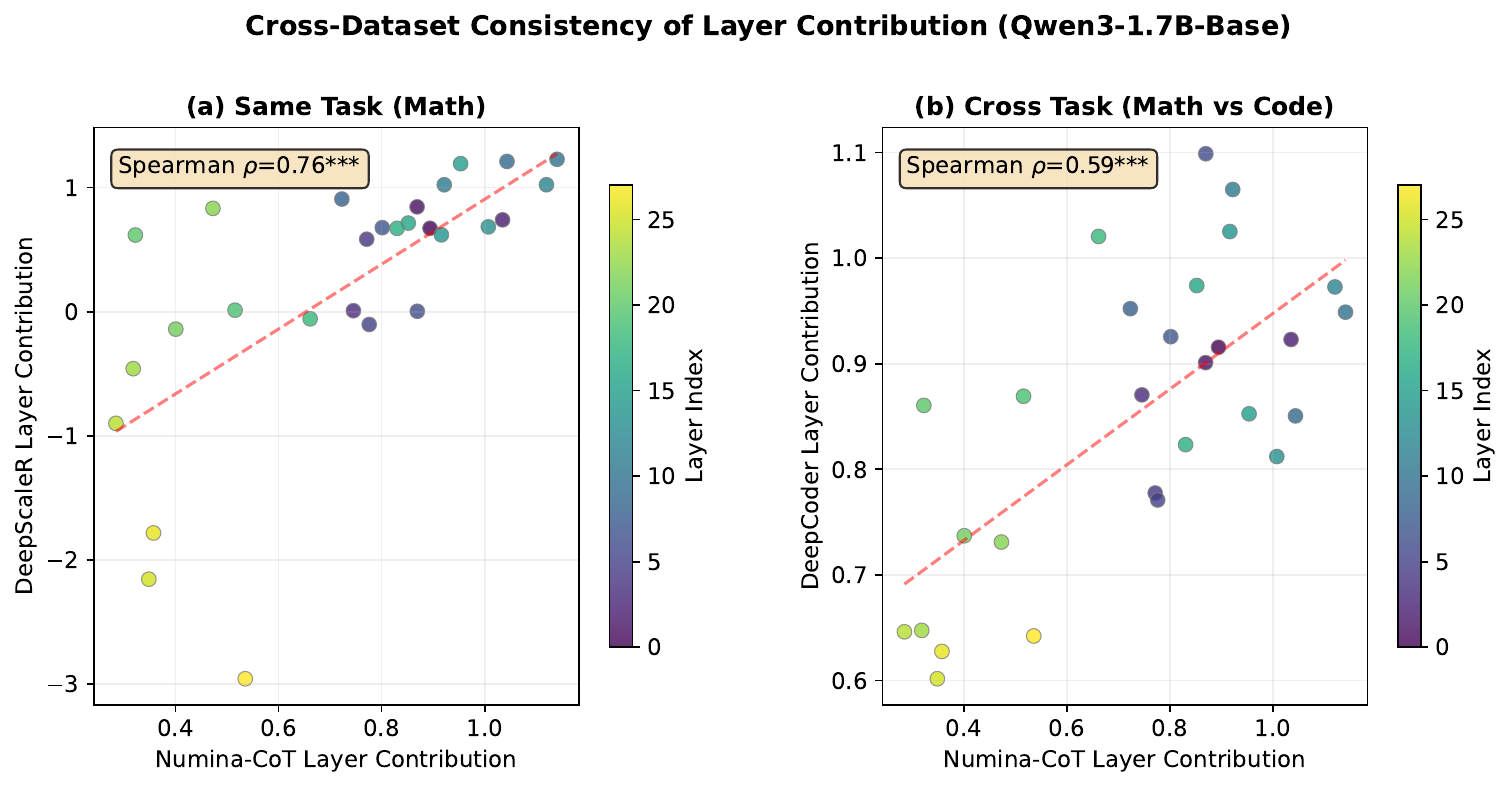}
    \caption{Cross-dataset consistency of layer contribution on
    Qwen3-1.7B-Base. Each point represents a single layer. (a)
    NuminaMath-CoT vs.\ DeepScaleR (both math). (b) NuminaMath-CoT (math)
    vs.\ DeepCoder (code).}
    \label{fig:cross_dataset}
\end{figure}

A natural question is whether the layer contribution patterns observed
in \S\ref{sec:contribution} are specific to the training dataset, or
reflect a more fundamental property of the model. To test this, we
repeat our single-layer training experiments on Qwen3-1.7B-Base using
two additional datasets: DeepScaleR~\citep{deepscaler2025}, a
mathematics dataset, and DeepCoder~\citep{deepcoder2025}, a coding
dataset.

We first compare layer contribution across two math datasets:
NuminaMath-CoT and DeepScaleR. For each dataset, we compute the layer
contribution $\mathcal{C}(k)$ for all 28 layers and rank them
accordingly. We then measure the consistency between the two rankings
using the Spearman rank correlation coefficient, which captures
whether the relative ordering of layers is preserved regardless of
differences in absolute contribution values. Despite differences in
data composition and difficulty, the per-layer contribution rankings
are strongly correlated (Spearman $\rho=0.76$, $p<0.001$).
Figure~\ref{fig:cross_dataset}(a) visualizes this correspondence:
each point represents a single layer, and layers that rank highly
under one dataset consistently rank highly under the other. This
suggests that layer contribution is not driven by the specific
content of the training data, but by the model's internal structure.

We further test whether this consistency extends across tasks by
comparing NuminaMath-CoT (math) and DeepCoder (code), which target
fundamentally different capabilities. The per-layer
rankings remain correlated (Spearman $\rho=0.59$, $p<0.001$;
Figure~\ref{fig:cross_dataset}(b)), indicating that even when the
training objective changes from mathematical reasoning to code
generation, the same layers tend to have the highest contribution.

Taken together, these results establish that layer contribution is an
intrinsic property of the pretrained model, determined by its
pretrained weights rather than the specific training data or task.
This has a direct practical implication: layer selections derived from
a smaller or more accessible dataset can be reliably transferred to
guide training on other data, a possibility we explore in
\S\ref{sec:guiding}.

\subsection{Generalization Across Model Families, Algorithms, and Tasks}
\label{sec:generalization_family}

\begin{figure}[t]
    \centering
    \includegraphics[width=0.62\linewidth]{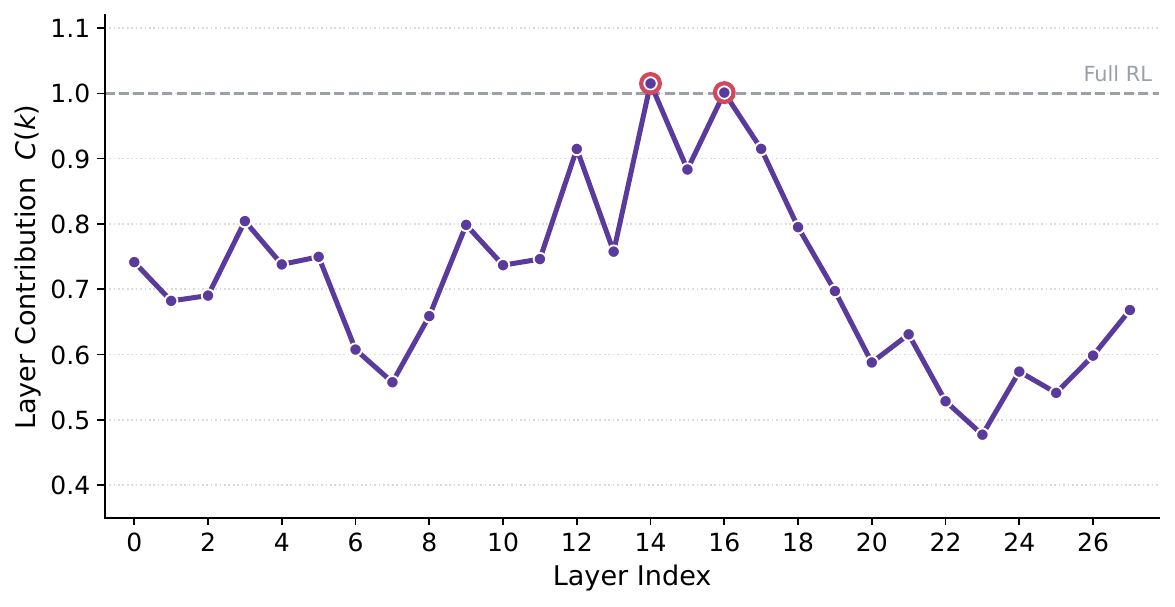}
    \caption{Layer contribution $C(k)$ for Qwen2.5-Math-1.5B (28 layers) trained with
    Dr.\,GRPO. Each point corresponds to one transformer layer trained in isolation. The
    dashed line marks full-parameter training ($C=1.0$); circled markers indicate layers
    that reach or exceed it. Despite the change in both model family and RL algorithm, the
    contribution profile retains the same structure observed on Qwen3: middle layers
    contribute most, while layers near the input and output ends contribute less.}
    \label{fig:contribution_qwen25math}
\end{figure}

\begin{table}[t]
\centering
\caption{Per-layer training results on Qwen2.5-Math-1.5B (Dr.\,GRPO). We report each math
benchmark and the overall average over the six benchmarks (Avg). $\mathcal{C}$ denotes
layer contribution computed on Avg. Complete per-layer results are in
Appendix~\ref{app:full_results}. $^\dagger$Official result from~\citet{liu2025understanding}.}
\label{tab:qwen25math_main}
\begin{tabular}{lcccccccc}
\toprule
Setting & AIME & AIME25 & AMC & MATH500 & Minerva & Olymp. & Avg & $\mathcal{C}$ \\
\midrule
Base    & 20.0 & 6.7  & 32.5 & 33.0 & 12.5 & 22.8 & 21.2 & 0.00 \\
Full    & 16.7 & 10.0 & 51.8 & 74.4 & 25.0 & 38.8 & 36.1 & 1.00 \\
Dr.\,GRPO$^\dagger$ & 20.0 & 6.7 & 53.0 & 74.2 & 25.7 & 37.6 & 36.2 & -- \\
\midrule
Layer 14 & 20.0 & 10.0 & 52.3 & 74.8 & 25.6 & 35.3 & \textbf{36.3} & \textbf{1.01} \\
Layer 16 & 20.0 & 10.0 & 51.8 & 75.2 & 24.9 & 34.9 & 36.1 & 1.00 \\
Layer 12 & 20.0 & 10.0 & 45.8 & 73.8 & 25.0 & 34.8 & 34.9 & 0.92 \\
Layer 8  & 13.3 & 3.3  & 43.4 & 69.4 & 20.6 & 30.7 & 30.1 & 0.60 \\
Layer 23 & 10.0 & 3.3  & 38.6 & 64.0 & 19.9 & 29.3 & 27.5 & 0.42 \\
\bottomrule
\end{tabular}
\end{table}

{The consistency observed in \S\ref{sec:consistency} shows that layer contribution is robust to changes in the training data and task within a single model. A stronger test is whether the phenomenon persists when the model family, the RL algorithm, or the task domain changes. We examine each of these axes below. 

\paragraph{Different model family and RL algorithm (Qwen2.5-Math-1.5B, Dr.~GRPO).}
We first test whether the findings transfer when both the model family and the RL algorithm change simultaneously. We repeat our single-layer training experiments on Qwen2.5-Math-1.5B, a model from a different family and pretraining recipe than the Qwen3 models studied above, and replace GRPO with Dr.~GRPO~\citep{liu2025understanding} as the optimization algorithm, following its training recipe (full details in Appendix~\ref{app:drgrpo_setup}). This setup differs from our main experiments along two independent axes.

Figure~\ref{fig:contribution_qwen25math} and Table~\ref{tab:qwen25math_main} show the results. The qualitative structure is identical to that of the Qwen3 models in \S\ref{sec:contribution}: layer contribution rises toward the middle of the network and falls off near both ends. The highest-contribution layers are concentrated in the middle of the stack (Layer~14, $\mathcal{C}=1.01$; Layer~16, $\mathcal{C}=1.00$; Layer~12, $\mathcal{C}=0.92$; Layer~15, $\mathcal{C}=0.89$; Layer~17, $\mathcal{C}=0.88$), whereas the lowest-contribution layers lie toward the later part of the network (Layer~23, $\mathcal{C}=0.42$; Layer~22, $\mathcal{C}=0.43$; Layer~25, $\mathcal{C}=0.48$). The best layer again recovers more than twice the contribution of the worst, reproducing the dramatic layer-wise variation seen on Qwen3. Moreover, a couple of middle layers (Layers~14 and 16) reach or slightly exceed the full-parameter baseline ($\mathcal{C} \geq 1.0$), echoing our earlier finding that single-layer training can match or surpass full-parameter RL.

\paragraph{Different task domain: agentic tasks (Qwen2.5-Instruct, GiGPO, ALFWorld).}

\begin{figure}[t]
    \centering
    \includegraphics[width=\textwidth]{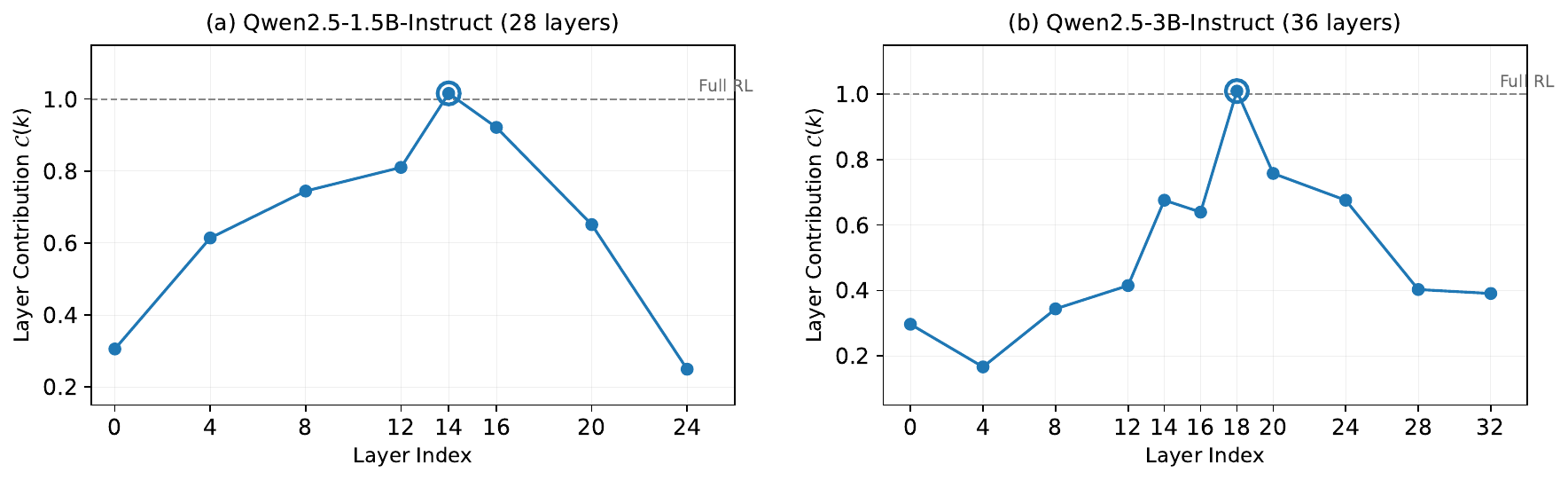}
    \caption{Layer contribution $\mathcal{C}(k)$ on the agentic task ALFWorld, trained with GiGPO. (a)~Qwen2.5-1.5B-Instruct (28 layers). (b)~Qwen2.5-3B-Instruct (36 layers). A representative subset of layers is trained due to computational constraints. The dashed line marks full-parameter training ($\mathcal{C}=1.0$); circled markers indicate layers that reach or exceed it. Despite the shift from mathematical reasoning to multi-step agentic decision-making, and the change in model family and RL algorithm, the contribution profile retains the same middle-layer concentration structure: Layer~14 ($\mathcal{C}=1.02$) on the 1.5B model and Layer~18 ($\mathcal{C}=1.01$) on the 3B model each surpass full-parameter training, while early and late layers contribute substantially less. }
    \label{fig:contribution_alfworld}
\end{figure}

\begin{table}[t]
\centering
\caption{Per-layer training results on Qwen2.5-1.5B-Instruct (GiGPO, ALFWorld). We report success rate (\%) on each ALFWorld task category and the overall average. $\mathcal{C}$ denotes layer contribution computed on the overall score. A representative subset of layers is trained. $^\dagger$Official result from~\citet{feng2025groupingrouppolicyoptimizationllm}.}
\label{tab:alfworld_15b}
\vspace{2mm}
\resizebox{\textwidth}{!}{%
\begin{tabular}{lcccccccc}
\toprule
Setting & Pick\&Place & Pick2\&Place & LookInLight & Heat\&Place & Cool\&Place & Clean\&Place & Overall & $\mathcal{C}$ \\
\midrule
Base     & 5.9  & 0.0  & 5.5  & 9.7  & 4.2  & 3.3  & 4.1  & 0.00 \\
Full     & 100  & 81.0 & 91.7 & 83.3 & 81.8 & 88.9 & 87.8 & 1.00 \\
GiGPO$^\dagger$ & 94.4 & 76.4 & 67.5 & 94.4 & 79.8 & 94.8 & 86.7 & -- \\
\midrule
Layer 14 & 100  & 85.7 & 100  & 83.3 & 81.8 & 77.8 & \textbf{89.1} & \textbf{1.02} \\
Layer 16 & 91.9 & 52.4 & 91.7 & 94.4 & 72.7 & 83.3 & 81.2 & 0.92 \\
Layer 12 & 83.8 & 47.6 & 66.7 & 72.2 & 81.8 & 66.7 & 71.9 & 0.81 \\
Layer 8  & 75.7 & 47.6 & 66.7 & 66.7 & 63.6 & 72.2 & 66.4 & 0.74 \\
Layer 20 & 67.6 & 47.6 & 66.7 & 50.0 & 50.0 & 66.7 & 58.6 & 0.65 \\
Layer 4  & 59.5 & 42.9 & 75.0 & 55.6 & 50.0 & 55.6 & 55.5 & 0.61 \\
Layer 0  & 48.6 & 23.8 & 41.7 & 16.7 & 13.6 & 22.2 & 29.7 & 0.31 \\
Layer 24 & 32.4 & 19.0 & 41.7 & 11.1 & 18.2 & 27.8 & 25.0 & 0.25 \\
\bottomrule
\end{tabular}
}
\end{table}

\begin{table}[t]
\centering
\caption{Per-layer training results on Qwen2.5-3B-Instruct (GiGPO, ALFWorld). We report success rate (\%) on each ALFWorld task category and the overall average. $\mathcal{C}$ denotes layer contribution computed on the overall score. A representative subset of layers is trained.}
\label{tab:alfworld_3b}
\vspace{2mm}
\resizebox{\textwidth}{!}{%
\begin{tabular}{lcccccccc}
\toprule
Setting & Pick\&Place & Pick2\&Place & LookInLight & Heat\&Place & Cool\&Place & Clean\&Place & Overall & $\mathcal{C}$ \\
\midrule
Base     & 57.6 & 9.1  & 37.5 & 0.0  & 12.5 & 8.0  & 24.2 & 0.00 \\
Full     & 100  & 81.0 & 75.0 & 83.3 & 86.4 & 100  & 90.2 & 1.00 \\
\midrule
Layer 18 & 94.6 & 76.2 & 100  & 83.3 & 86.4 & 100  & \textbf{90.8} & \textbf{1.01} \\
Layer 20 & 97.3 & 52.4 & 100  & 33.3 & 77.3 & 72.2 & 74.2 & 0.76 \\
Layer 14 & 89.2 & 52.4 & 75.0 & 38.9 & 68.2 & 72.2 & 68.8 & 0.68 \\
Layer 24 & 97.3 & 47.6 & 58.3 & 55.6 & 54.5 & 72.2 & 68.8 & 0.68 \\
Layer 16 & 91.9 & 52.4 & 50.0 & 50.0 & 63.6 & 61.1 & 66.4 & 0.64 \\
Layer 12 & 78.4 & 28.6 & 66.7 & 44.4 & 40.9 & 33.3 & 51.6 & 0.42 \\
Layer 28 & 79.5 & 36.8 & 44.4 & 27.3 & 32.0 & 48.0 & 50.8 & 0.40 \\
Layer 32 & 81.1 & 42.9 & 50.0 & 22.2 & 36.4 & 38.9 & 50.0 & 0.39 \\
Layer 8  & 81.1 & 28.6 & 50.0 & 33.3 & 27.3 & 33.3 & 46.9 & 0.34 \\
Layer 0  & 78.4 & 28.6 & 41.7 & 38.9 & 18.2 & 27.8 & 43.8 & 0.30 \\
Layer 4  & 67.6 & 28.6 & 41.7 & 22.2 & 18.2 & 5.6  & 35.2 & 0.17 \\
\bottomrule
\end{tabular}
}
\end{table}

The experiments above focus on mathematical reasoning and code generation. To test whether layer contribution generalizes to a fundamentally different task domain, we conduct single-layer training on Qwen2.5-1.5B-Instruct (28 layers) and Qwen2.5-3B-Instruct (36 layers) using GiGPO on the agentic benchmark ALFWorld. Unlike the mathematical setting, agentic tasks require multi-step decision-making in interactive environments, representing a qualitatively different capability.

Despite the change in task domain, model family, and RL algorithm relative to our main Qwen3 experiments, the same structural pattern emerges (Figure~\ref{fig:contribution_alfworld}). On Qwen2.5-1.5B-Instruct, the best single layer (Layer~14) achieves $\mathcal{C}=1.02$, surpassing full-parameter training, while the weakest layer (Layer~24) reaches only $\mathcal{C}=0.25$ (Table~\ref{tab:alfworld_15b}). High-contribution layers again concentrate in the middle of the network. On Qwen2.5-3B-Instruct, the pattern is consistent: the best layer (Layer~18) reaches $\mathcal{C}=1.01$, again surpassing full-parameter training, while the weakest (Layer~4) reaches only $\mathcal{C}=0.17$ (Table~\ref{tab:alfworld_3b}). Notably, the RL gain on these agentic tasks is far larger than in the mathematical setting (83.7 and 66.0 points respectively, compared to 6--10 points on math), yet the middle-layer concentration structure persists, indicating that the pattern is not limited to small-magnitude adaptations. These results demonstrate that the layer contribution structure is not specific to reasoning or coding tasks but extends to agentic problem-solving.
\paragraph{Different model architecture (DeepSeek-Distilled-Qwen-7B, GRPO, Skywork).}

\begin{figure}[t]
    \centering
    \includegraphics[width=0.62\linewidth]{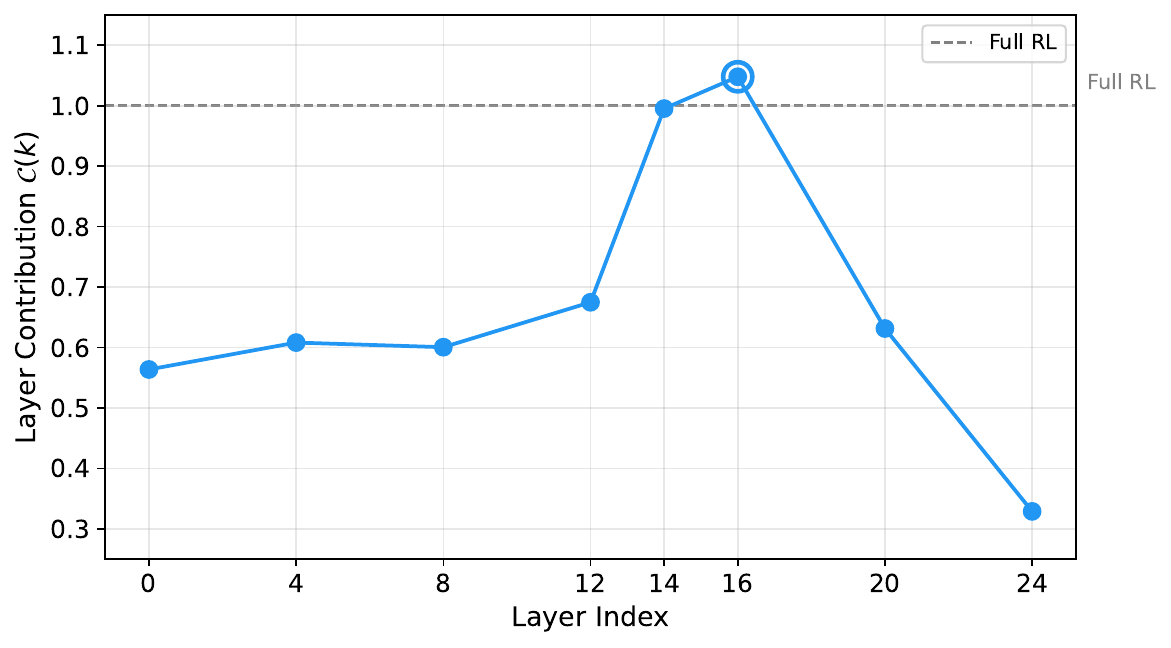}
    \caption{Layer contribution $\mathcal{C}(k)$ for DeepSeek-Distilled-Qwen-7B (28 layers) trained with GRPO on the Skywork mathematics dataset. Only a subset of layers (0, 4, 8, 12, 14, 16, 20, 24) are trained due to computational constraints. The dashed line marks full-parameter training ($\mathcal{C}=1.0$); the circled marker indicates the layer that exceeds it. Despite differing from the Qwen3 and Qwen2.5 models in both pretraining recipe (distilled from DeepSeek-R1) and training data, the contribution profile retains the same middle-layer concentration structure: Layer~16 ($\mathcal{C}=1.05$) surpasses full-parameter training, while early and late layers contribute substantially less.}
    \label{fig:contribution_deepseek7b}
\end{figure}

\begin{table}[t]
\centering
\caption{Per-layer training results on DeepSeek-Distilled-Qwen-7B (GRPO, Skywork). We report each math benchmark and the overall average over six benchmarks (Avg). $\mathcal{C}$ denotes layer contribution computed on Avg. Due to computational constraints, we train a representative subset of layers spanning the full depth of the network.}
\label{tab:deepseek7b_main}
\begin{tabular}{lcccccccc}
\toprule
Setting & AIME & AIME25 & AMC & MATH500 & Minerva & Olymp. & Avg & $\mathcal{C}$ \\
\midrule
Base    & 47.2 & 35.3 & 69.9 & 88.2 & 34.6 & 49.0 & 54.1 & 0.00 \\
Full    & 55.0 & 45.0 & 83.1 & 94.0 & 41.2 & 68.7 & 64.5 & 1.00 \\
\midrule
Layer 16 & 57.5 & 45.0 & 86.7 & 96.6 & 38.6 & 65.6 & \textbf{65.0} & \textbf{1.05} \\
Layer 14 & 55.0 & 38.3 & 86.7 & 95.6 & 43.4 & 67.7 & 64.5 & 1.00 \\
Layer 12 & 53.3 & 37.1 & 81.9 & 92.8 & 39.3 & 62.2 & 61.1 & 0.67 \\
Layer 20 & 57.9 & 36.7 & 77.1 & 93.0 & 40.4 & 58.8 & 60.6 & 0.63 \\
Layer 4  & 50.8 & 37.5 & 81.9 & 93.2 & 39.7 & 59.3 & 60.4 & 0.61 \\
Layer 8  & 51.2 & 37.5 & 79.5 & 91.4 & 42.3 & 60.0 & 60.3 & 0.60 \\
Layer 0  & 50.4 & 35.4 & 83.1 & 93.2 & 38.2 & 59.3 & 59.9 & 0.56 \\
Layer 24 & 52.5 & 35.4 & 73.5 & 90.8 & 37.6 & 55.1 & 57.5 & 0.33 \\
\bottomrule
\end{tabular}
\end{table}

To further test generality beyond the Qwen3 model families, we conduct partial-layer experiments on DeepSeek-Distilled-Qwen-7B (28 layers), a model distilled from DeepSeek-R1 into the Qwen architecture. We train with GRPO on the Skywork~\citep{he2025skyworkopenreasoner1} mathematics dataset. 

The same structure reappears: middle layers exhibit the highest contribution, with the best layer reaching $\mathcal{C}=\text{1.05}$ and the weakest reaching $\mathcal{C}=\text{0.33}$. The best single layer again surpasses full-parameter training. This confirms that the phenomenon is not confined to models pretrained from scratch but also holds for distilled models.

\paragraph{Summary across all seven models.}
Table~\ref{tab:generalization_summary} summarizes the key layer contribution statistics across all seven models. Despite variation in model family, scale, RL algorithm, training dataset, and task domain, every model exhibits the same qualitative structure: (1) the best single layer matches or surpasses full-parameter training ($\mathcal{C} \geq 1.0$), (2) high-contribution layers concentrate in the middle of the network, and (3) layers near the input and output ends contribute substantially less.

\begin{table}[t]
\centering
\caption{Layer contribution summary across all seven models. For each model we report the best and worst layer contribution, the number of layers with $\mathcal{C} \geq 1.0$, and whether the contribution profile exhibits a middle-layer concentration shape. All models show the same qualitative pattern. $^\dagger$Only a representative subset of layers is trained.}
\label{tab:generalization_summary}
\vspace{2mm}
\resizebox{\textwidth}{!}{%
\begin{tabular}{llllcccc}
\toprule
\textbf{Model} & \textbf{Family} & \textbf{Algorithm} & \textbf{Task} & \textbf{Best $\mathcal{C}$} & \textbf{Worst $\mathcal{C}$} & \textbf{Layers $\geq 1.0$} & \textbf{Middle-layer concentration} \\
\midrule
Qwen3-1.7B-Base & Qwen3 & GRPO & Math & 1.14 & 0.28 & 5/28 & \checkmark \\
Qwen3-4B-Base & Qwen3 & GRPO & Math & 1.06 & 0.66 & 4/36 & \checkmark \\
Qwen3-8B-Base & Qwen3 & GRPO & Math & 1.07 & $-$0.51 & 4/36 & \checkmark \\
Qwen2.5-Math-1.5B & Qwen2.5 & Dr.~GRPO & Math & 1.01 & 0.42 & 2/28 & \checkmark \\
Qwen2.5-1.5B-Instruct & Qwen2.5 & GiGPO & Agentic & 1.02 & 0.25 & 1/8$^\dagger$ & \checkmark \\
Qwen2.5-3B-Instruct & Qwen2.5 & GiGPO & Agentic & 1.01 & 0.17 & 1/11$^\dagger$ & \checkmark \\
DeepSeek-Distilled-Qwen-7B & Qwen2.5 & GRPO & Math & 1.05 & 0.33 & 2/8$^\dagger$ & \checkmark \\
\bottomrule
\end{tabular}
}
\end{table}

These results, together with the cross-dataset and cross-task consistency established in \S\ref{sec:consistency}, demonstrate that the observed layer contribution pattern is remarkably stable. Across seven models spanning two model families, three RL algorithms, multiple datasets, three task domains, and model scales from 1.5B to 8B, the same qualitative behavior consistently emerges: RL gains are highly uneven across transformer layers, concentrate in the middle of the network, and can often be largely recovered by training only a single layer. Collectively, these findings suggest that the gains from RL post-training emerge primarily through the adaptation of a relatively small subset of transformer layers, rather than through uniform adaptation across the entire network.

\section{Guiding Full-Parameter RLVR by Layer Contribution}
\label{sec:guiding}

The experiments in \S\ref{sec:experiments} establish that layers
differ dramatically in contribution.  This raises a practical
question: \textit{Can this observation be used to improve standard
full-parameter RLVR, which treats all layers uniformly during
training?} Since different layers vary in their capacity to absorb RL
training signals, differentiating across layers according to their
contribution should yield better outcomes than uniform treatment. We
explore three strategies: adjusting per-layer learning rates based on
layer contribution (\S\ref{sec:adaptive_lr}), selectively training
only the highest-contribution layers (\S\ref{sec:selective}), and a heuristic method for selective training based on layer position
(\S\ref{sec:heuristic}). All experiments in this section use
NuminaMath-CoT as the training dataset and report math performance
averaged over the same four in-domain benchmarks (MATH500, GSM8K,
OlympiadBench, AMC) as in \S\ref{sec:experiments}, unless otherwise
noted. All results are reported as mean $\pm$ standard deviation over
3 independent evaluation runs.

\begin{figure}[t]
    \centering
    \includegraphics[width=\textwidth]{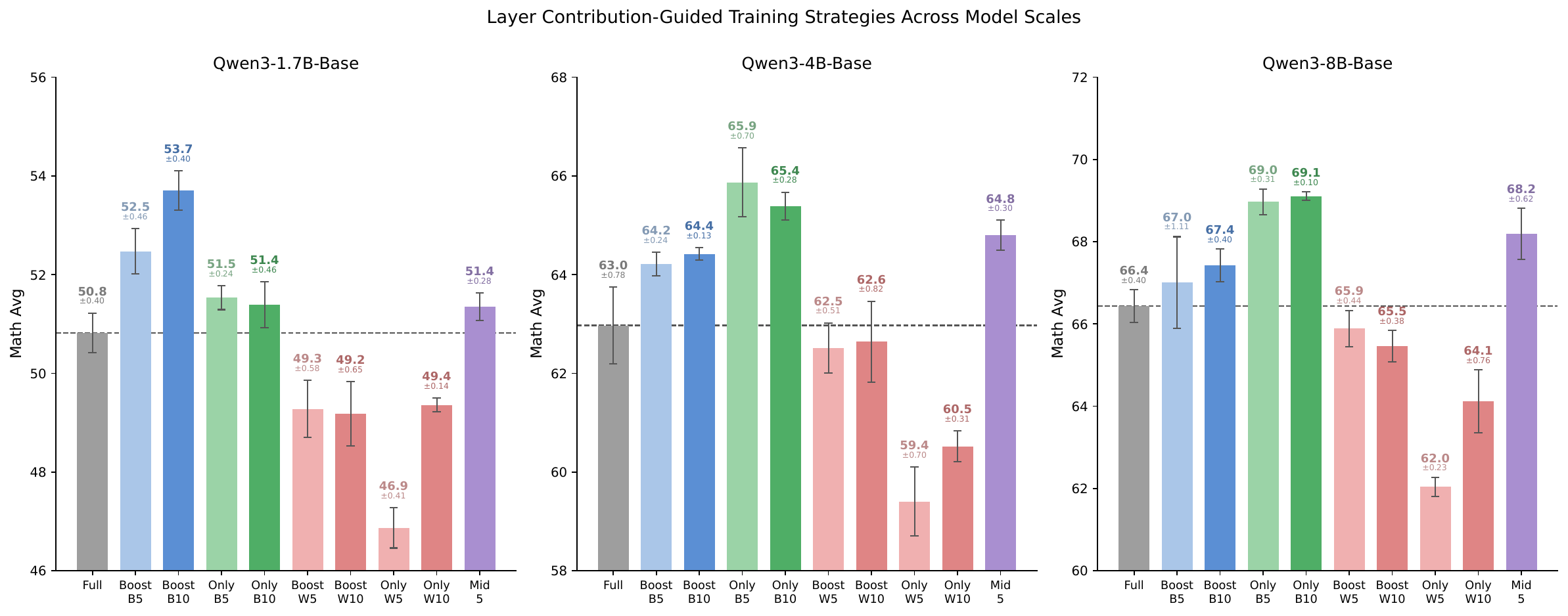}
    \caption{Layer contribution-guided training strategies across model
    scales. {\color{blue}Blue}: boosting high-contribution layers with increased
    learning rate. {\color{green}Green}: training only high-contribution layers.
    {\color{red}Red}: control experiments using low-contribution layers. {\color{violet}Purple}:
    position-based middle-layer heuristic. {\color{gray}Grey}: full-parameter baseline
    (also indicated by dashed line). B$k$/W$k$: Best/Worst
    $k$ layers by layer contribution. Error bars denote one standard
    deviation over 3 independent training runs.}
    \label{fig:guided_training}
\end{figure}

\subsection{Layer-Adaptive Learning Rate}
\label{sec:adaptive_lr}

Standard full-parameter RLVR applies a uniform learning rate across
all layers. However, since layers differ substantially in their
capacity to learn from RL signals, a uniform learning rate may be
suboptimal. We therefore experiment with assigning higher learning
rates to high-contribution layers while keeping other layers at the
base learning rate.

Specifically, we select the best $k$ layers ranked by layer
contribution (denoted B$k$, e.g., B5 = the 5 highest-contribution
layers) and increase their learning rate to~$1 \times 10^{-5}$, while all remaining layers
are trained at the default rate~$5 \times 10^{-6}$. As a control experiment, we also boost the
worst $k$ layers (denoted W$k$). We experiment with $k \in \{5, 10\}$
across all three model scales (Figure~\ref{fig:guided_training}).
Across models and configurations, boosting high-contribution layers
consistently improves math performance over the uniform-lr baseline:
on Qwen3-1.7B-Base, Boost B10 achieves $53.70 \pm 0.40$ compared to the
$50.82 \pm 0.40$ baseline (+2.88, representing 43\% of the total RL
training gain); on Qwen3-4B-Base, our B10 gets $64.42 \pm 0.13$ vs.\ $62.97 \pm 0.78$
(+1.45); and on Qwen3-8B-Base, Boost B10 reaches $67.42 \pm 0.40$ vs.\
$66.43 \pm 0.40$ (+0.99). In contrast, boosting the
lowest-contribution layers (Boost W$k$) leads to a decline in
performance across all three models. This asymmetry confirms that the
improvement is driven by the contribution-guided selection rather than
the learning rate adjustment itself.

\subsection{Layer-Selective Training}
\label{sec:selective}

We further ask whether low-contribution layers can be frozen entirely
during training. We train only the best $k$ layers while keeping all
remaining layers frozen, with $k \in \{5, 10\}$ across all three
model scales (Figure~\ref{fig:guided_training}).

On Qwen3-1.7B-Base, training only the best layers already exceeds the
full-parameter baseline (Only B5: $51.53 \pm 0.24$, Only B10:
$51.39 \pm 0.46$, vs.\ baseline $50.82 \pm 0.40$). On larger models, the
improvement is more pronounced: on Qwen3-4B-Base, Only B5 reaches
$65.87 \pm 0.70$ (+2.90 over baseline, representing 27\% of the total
RL gain), and on Qwen3-8B-Base, Only B10 reaches $69.11 \pm 0.10$
(+2.68, representing 32\% of the total RL gain). In both cases,
selective training surpasses not only full-parameter training but also
the adaptive learning rate strategy from \S\ref{sec:adaptive_lr}. This
suggests that at larger scales, updates to low-contribution layers may
not contribute positively to training, and freezing them yields a
cleaner optimization. Conversely, training only the worst $k$ layers
leads to substantially worse results across all scales (e.g., Only W5:
$46.87 \pm 0.41$ on 1.7B, $59.40 \pm 0.70$ on 4B, $62.04 \pm 0.23$ on
8B), confirming that the effective learning in RLVR is concentrated in
high-contribution layers.

\subsection{Heuristic Layer Selection}
\label{sec:heuristic}

The preceding strategies require layer contribution rankings derived
from per-layer training, which is expensive and impractical for routine use. We explore whether this profiling step can be
bypassed altogether. Since layer contribution consistently exhibits a
pattern of higher values in middle layers and lower values near the
input and output ends across all three model scales
(Figure~\ref{fig:contribution}), we test a simple heuristic: select
the middle $k$ layers by position, without any profiling at all.

Specifically, for a model with $L$ layers, we select layers in the
range $[\lfloor L/2 - k/2 \rfloor, \lfloor L/2 + k/2 \rfloor)$ and
apply the same selective training setup as in
\S\ref{sec:selective}. Setting $k=5$ yields layers 11--15 for
Qwen3-1.7B-Base ($L=28$) and layers 15--19 for both Qwen3-4B-Base and
Qwen3-8B-Base ($L=36$). On Qwen3-1.7B-Base, heuristic selection of
the middle 5 layers achieves a math performance of $51.35 \pm 0.28$,
compared to $51.53 \pm 0.24$ for contribution-guided selection (Only
B5) and $50.82 \pm 0.40$ for the full-parameter baseline (+0.53 over
Full, capturing 8\% of the total RL training gain). On Qwen3-4B-Base,
the heuristic achieves $64.80 \pm 0.30$ vs.\ $65.87 \pm 0.70$ (Only B5)
and $62.97 \pm 0.78$ (Full), a +1.83 improvement representing 17\% of
the total RL gain. On Qwen3-8B-Base, $68.19 \pm 0.62$ vs.\
$68.97 \pm 0.31$ (Only B5) and $66.43 \pm 0.40$ (Full), a +1.76
improvement representing 21\% of the total RL gain. Across all three
scales, the heuristic surpasses the full-parameter baseline without any
per-layer profiling, and achieves a substantial portion of the
improvement of contribution-guided selection.

This result has a practical implication: when even a single round of
per-layer profiling is unavailable, simply training the middle layers
provides a strong default strategy that captures a meaningful portion
of the benefit of contribution-guided selection.

\section{Discussion}
\label{sec:discussion}

\subsection{Diversity in Layer-Trained Models}
\label{sec:diverse_experts}

\begin{figure}[t]
    \centering
    \includegraphics[width=0.55\linewidth]{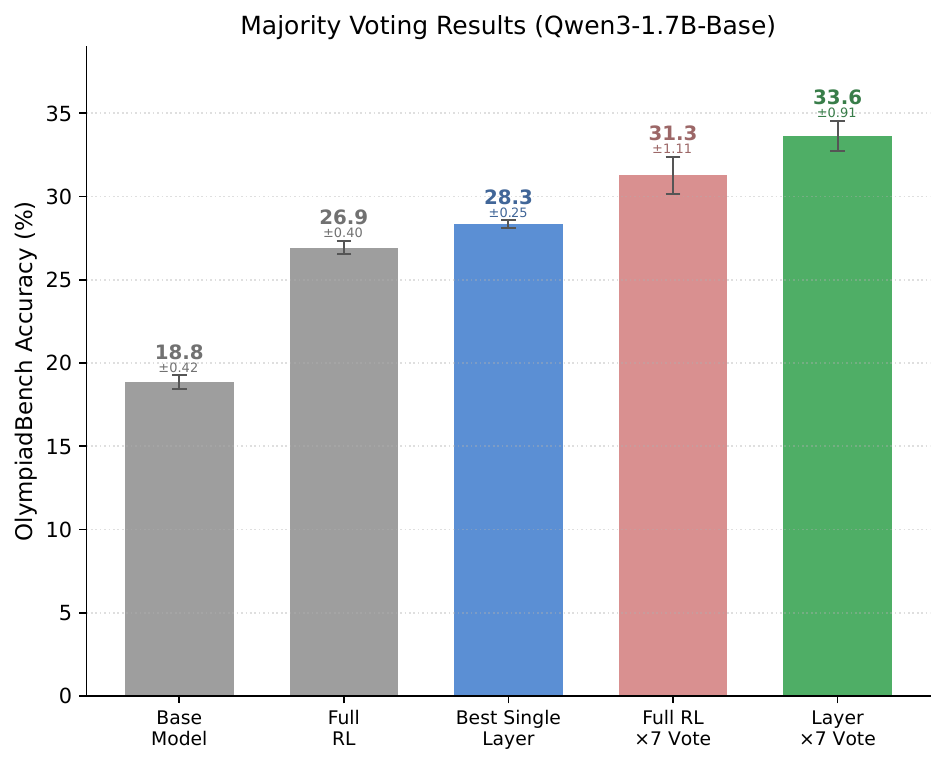}
    \caption{Majority voting results on OlympiadBench (Qwen3-1.7B-Base). Voting across $7$
    diverse layer-trained models surpasses both the best single-layer model and the
    full-parameter baseline, and also exceeds self-consistency--style voting over $7$ samples
    from the same full-parameter model. Error bars denote one standard deviation over $3$
    independent evaluation runs.}
    \label{fig:majority_voting}
\end{figure}

In \S\ref{sec:experiments} we showed that layer contribution varies dramatically across layers. Among the high-contribution layers, however, some layer-trained models' performance is comparable. For example, on Qwen3-1.7B-Base, the top-7 layers all achieve OlympiadBench accuracy between $23\%$ and $28\%$. A natural question is whether these similarly-performing models solve the same problems or different ones. We investigate this below to better understand what has been learned in layer training.

\paragraph{High-contribution layers solve different subsets of problems.}
We train each of the 28 layers of Qwen3-1.7B-Base independently (as in \S\ref{sec:contribution}) and evaluate all resulting models on OlympiadBench. For each of the top-7 high-contribution layers, we record the set of problems its model newly solves relative to the base model, and measure pairwise overlap via the Jaccard similarity $J(A,B) = |A \cap B| / |A \cup B|$, where A and B are problems solved by two different layer-trained models. The average pairwise Jaccard similarity is only $34.1\%$: despite comparable accuracy, these models solve largely non-overlapping sets of problems. For instance, the Layer~10 and Layer~13 models reach $27.3\%$ and $28.3\%$ accuracy respectively, but share only $31.9\%$ of their newly-solved problems. This indicates that each high-contribution layer captures a different aspect of RL-induced improvement, and the knowledge they encode is complementary rather than redundant.

\paragraph{Majority voting quantifies this complementarity.}
To measure the practical value of this complementarity, we aggregate the predictions of the top-7 layer-trained models via majority voting on OlympiadBench. The ensemble reaches $33.6 \pm 0.91\%$, surpassing both the best individual layer-trained model ($28.3 \pm 0.25\%$) and the full-parameter baseline ($26.9 \pm 0.40\%$) by a substantial margin (Figure~\ref{fig:majority_voting}). To isolate the role of \emph{structural} diversity from sampling randomness, we compare against self-consistency--style voting~\citep{wang2023selfconsistencyimproveschainthought}, which aggregates $7$ independent samples from the same full-parameter model; this reaches only $31.3 \pm 1.11\%$, confirming that diversity arising from training distinct layers is more effective than diversity from repeated sampling. This connects to the observation by \citet{gan2026neuralthicketsdiversetask} that diverse task-specific experts are densely distributed around pretrained weights: single-layer training provides a structured way to reach such experts along an interpretable axis---transformer depth.

Finally, we emphasize that the layer-wise training and majority voting framework is intended as an analysis tool rather than a practical training strategy, since it requires independently training multiple layer-specific models. Nevertheless, this analysis provides valuable insights into the diversity of representations learned by different transformer layers and their respective contributions to RL post-training. These insights, in turn, motivate the more practical layer-aware training strategies developed in Section~\ref{sec:guiding}.

\subsection{Weight Change Magnitude vs.\ Layer Contribution}
\label{sec:weight_change}

A natural question is whether high-contribution layers are simply layers whose parameters change more during the full-parameter training. For example, in the extreme case if after full training, only the middle layer moves while the rest of the layers stay unchanged, then of course our observation of middle-layer contribution is trivial.  To investigate this, we measure the per-layer weight change magnitude $\|\Delta\theta_k\|_2$ (L2 norm of the difference between trained and base-model parameters) for both full-parameter training and several representative single-layer training runs on Qwen3-1.7B-Base (Figure~\ref{fig:weight_change}).

Two observations stand out. First, under full-parameter training, the weight change magnitude is remarkably uniform across layers (ranging from approximately 0.5 to 0.8), despite the highly non-uniform layer contribution profile observed in \S\ref{sec:contribution}. This dissociation indicates that layer contribution is not explained by how much a layer's parameters change during training, as middle layers do not change more than other layers in full training, yet they contribute far more when trained in isolation.

Second, under single-layer training, the trained layer undergoes a substantially larger weight change than the same layer experiences during full training (approximately 0.8--1.0 vs.\ 0.5--0.7). This suggests that when a single layer must absorb all RL-induced improvement, it compensates by moving further in parameter space. Crucially, however, the magnitude of this single-layer weight change is similar across layers with very different contribution values: both high-contribution and low-contribution layers undergo comparable weight changes when trained alone, yet produce vastly different performance outcomes. This reinforces the conclusion that layer contribution reflects the \emph{effectiveness} of a layer's parameter subspace for capturing RL improvement, rather than the \emph{magnitude} of parameter change.

\begin{figure}[t]
    \centering
    \includegraphics[width=0.85\textwidth]{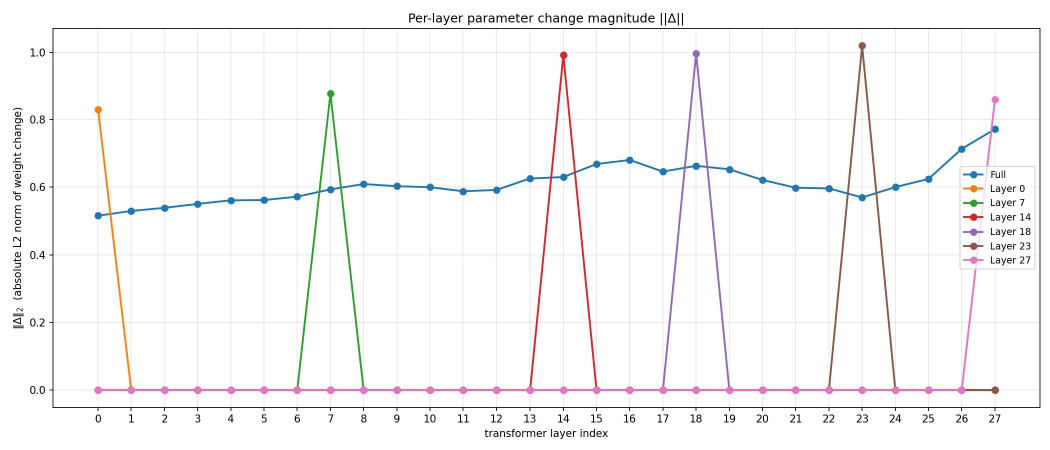}
    \caption{Per-layer weight change magnitude $\|\Delta\theta_k\|_2$ on Qwen3-1.7B-Base. Blue: full-parameter training (all layers change). Colored spikes: single-layer training (only the trained layer changes; all others remain at zero). Under full training, the weight change is relatively uniform across layers, contrasting with the highly non-uniform layer contribution profile. Under single-layer training, all trained layers undergo comparable weight changes regardless of their contribution, indicating that layer contribution reflects the effectiveness of a layer's parameter subspace rather than the magnitude of parameter change.}
    \label{fig:weight_change}
\end{figure}

\section{Related Work} \label{sec:related}

\paragraph{Layer Importance in LLMs.}
Several works have studied the roles of individual layers in LLMs.
\citet{song2026demystifyingrolesllmlayers} analyze layer importance
through pruning, finding that removing certain layers causes
performance to collapse while removing others has little effect.
\citet{zhang2024investigatinglayerimportancelarge} identify
``cornerstone layers'' whose removal reduces performance to near
random guessing, and \citet{nepal2025layerimportancemathematicalreasoning}
show that the critical layers for mathematical reasoning are
determined during pretraining and remain invariant after post-training.
In the context of parameter-efficient finetuning,
LISA~\citep{pan2024lisalayerwiseimportancesampling} proposes randomly
sampling layers to update during SFT, and
MISA~\citep{liu2026misamemoryefficientllmsoptimization} extends this
with importance-aware sampling.
AdaGradSelect~\citep{kumar2025adagradselectadaptivegradientguidedlayer}
uses gradient statistics to dynamically select which layers to train
during SFT. These works primarily operate in inference-time analysis
or the supervised finetuning setting. In contrast, our work focuses on
the RL setting and demonstrates that layer contribution is a stable
property that generalizes across datasets and tasks, enabling
principled layer selection without per-dataset profiling.

\paragraph{Diverse Solutions in Weight Space.}
\citet{gan2026neuralthicketsdiversetask} discover that the
neighborhood around pretrained weights contains dense, diverse
task-specific experts that can be found through random perturbation.
Our findings provide a new perspective: single-layer training
offers a structured way to explore this neighborhood, with each layer
accessing a different region of the solution space. The diversity
among layer-trained models is evidenced by their low pairwise agreement
and effective majority voting.
\section{Conclusion}
\label{sec:conclusion}

In this work, we investigate how different layers in LLMs respond to RL post-training. Through systematic single-layer training experiments across seven models spanning two model families (Qwen3, Qwen2.5), three RL algorithms (GRPO, GiGPO, Dr.~GRPO), and two task domains (mathematical reasoning and agentic decision-making), we discover that layers differ dramatically in their capacity to capture RL-induced improvement. High-contribution layers consistently concentrate in the middle of the network, and this variation is a stable structural property determined by the model itself rather than by the specific dataset, task, or algorithm used for training.

We show that this property has direct practical value: prioritizing high-contribution layers through adaptive learning rates or selective training consistently outperforms standard uniform RL, and even a simple heuristic that trains only the middle layers surpasses full-parameter training without any per-layer profiling. Beyond guiding training, we find that high-contribution layers capture complementary aspects of RL-induced improvement, and that the magnitude of weight change does not explain the variation in layer contribution---layers that change equally in parameter space produce vastly different performance outcomes.

Our work has several limitations. Our guided training strategies are validated only on mathematical reasoning; extending them to coding and agentic tasks is left for future work. Additionally, our layer contribution metric is defined with respect to a specific training configuration, and a deeper theoretical understanding of why middle layers are disproportionately important for RL adaptation remains an open question.

\bibliographystyle{plainnat}
\bibliography{references}

\section*{Appendix}

\appendix

\appendix

\section{Training Details and Hyperparameters}
\label{app:details}

\subsection{Overview}

Table~\ref{tab:setup_summary} in the main text summarizes all seven models and their training configurations. Below we provide full hyperparameter details for each experimental setup. In all cases, single-layer training freezes every parameter except the target decoder layer (including the embedding layer and the language model head), and follows the same hyperparameters as the corresponding full-parameter baseline unless otherwise noted.

\subsection{Qwen3 Models (GRPO, NuminaMath-CoT)}
\label{app:qwen3_setup}

All Qwen3 experiments are conducted using the veRL framework with GRPO as the RL algorithm and AdamW as the optimizer. We use Qwen3-1.7B-Base (28 layers), Qwen3-4B-Base (36 layers), and Qwen3-8B-Base (36 layers) as base models.

For single-layer training, each layer is trained independently on NuminaMath-CoT (downsampled to 50K problems after decontamination). The full-parameter baseline uses identical hyperparameters but with all layers unfrozen. Table~\ref{tab:hp_qwen3} summarizes the key hyperparameters. For the full-parameter baselines, we tune the learning rate over $\{1 \times 10^{-6},\, 3 \times 10^{-6},\, 5 \times 10^{-6},\, 1 \times 10^{-5}\}$ and report the best result (with learning rate $= 5 \times 10^{-6}$);  For single-layer training, we use a learning rate of $5 \times 10^{-6}$, which aligns with the full-parameter training. For the adaptive learning rate experiments in \S\ref{sec:adaptive_lr}, the boosted learning rate is set to $1 \times 10^{-5}$ and the base rate to $5 \times 10^{-6}$.

\begin{table}[h]
\centering
\caption{Training hyperparameters for Qwen3 models (GRPO, NuminaMath-CoT).}
\label{tab:hp_qwen3}
\vspace{2mm}
\begin{tabular}{lc}
\toprule
Hyperparameter & Value \\
\midrule
RL Algorithm & GRPO \\
Learning Rate & $5 \times 10^{-6}$ \\
Train Batch Size & 512 \\
PPO Mini Batch Size & 128 \\
PPO Micro Batch Size & 8 \\
Group Size ($G$) & 4 \\
Max Response Length & 3072 \\
KL Coefficient & 0.001 \\
Clip Range ($\epsilon$) & 0.2 \\
Epochs & 4 \\
\bottomrule
\end{tabular}
\end{table}
\subsection{Qwen2.5-Math-1.5B (Dr.\,GRPO)}
\label{app:drgrpo_setup}

The generalization experiment in \S\ref{sec:generalization_family} follows the Dr.\,GRPO recipe of \citet{liu2025understanding}, implemented in the Oat framework. We RL-tune Qwen2.5-Math-1.5B with the unbiased Dr.\,GRPO objective, which removes the response-length ($1/|o_i|$) and question-level (std) normalization terms from the GRPO objective. Training uses the MATH training set with a binary answer-matching reward verified by Math-Verify. The single-layer training protocol is identical to \S\ref{sec:contribution}: we train one decoder layer at a time while freezing all other parameters, and for each layer we report the best checkpoint by average score. Table~\ref{tab:hp_drgrpo} lists the hyperparameters.
Similar to the Qwen3 experiments, for the full-parameter baselines, we tune the learning rate over $\{1 \times 10^{-6},\, 3 \times 10^{-6},\, 5 \times 10^{-6},\, 1 \times 10^{-5}\}$ and report the best result (with learning rate $= 5 \times 10^{-6}$);  For single-layer training, we use a learning rate of $5 \times 10^{-6}$, which aligns with the full-parameter training.

For evaluation we use six mathematics benchmarks---AIME~2024, AIME~2025, AMC, MATH500, Minerva~Math, and OlympiadBench---and report layer contribution $\mathcal{C}(k)$ on their unweighted average (Avg6). This evaluation suite follows the setup associated with the Dr.\,GRPO recipe and differs from the benchmark suite used for the Qwen3 experiments.

\begin{table}[h]
\centering
\caption{Training hyperparameters for Qwen2.5-Math-1.5B (Dr.\,GRPO).}
\label{tab:hp_drgrpo}
\vspace{2mm}
\begin{tabular}{lc}
\toprule
Hyperparameter & Value \\
\midrule
RL Algorithm & Dr.\,GRPO \\
Learning Rate & $5 \times 10^{-6}$ \\
Train Batch Size & 128 \\
Group Size ($G$) & 8 \\
Max Response Length & 3072 \\
KL Coefficient & 0 \\
Clip Range ($\epsilon$) & 0.2 \\
Epochs & 8 \\
\bottomrule
\end{tabular}
\end{table}

\subsection{Qwen2.5-Instruct Models (GiGPO, ALFWorld)}
\label{app:gigpo_setup}

The agentic experiments in \S\ref{sec:generalization_family} use Qwen2.5-1.5B-Instruct (28 layers) and Qwen2.5-3B-Instruct (36 layers) trained with GiGPO on ALFWorld, an environment-based agentic benchmark consisting of 2,435 household tasks across six categories (Pick\&Place, Pick\&Place with two objects, Examine in Light, Heat\&Place, Cool\&Place, Clean\&Place). The reward is binary: 1 if the agent successfully completes the task, 0 otherwise. Table~\ref{tab:hp_gigpo} lists the hyperparameters.
Similar to the Qwen3 experiments, for the full-parameter baselines, we tune the learning rate over $\{1 \times 10^{-6},\, 3 \times 10^{-6},\, 5 \times 10^{-6},\, 1 \times 10^{-5}\}$ and report the best result (with learning rate $= 5 \times 10^{-6}$);  For single-layer training, we use a learning rate of $5 \times 10^{-6}$, which aligns with the full-parameter training.

\begin{table}[h]
\centering
\caption{Training hyperparameters for Qwen2.5-Instruct models (GiGPO, ALFWorld).}
\label{tab:hp_gigpo}
\vspace{2mm}
\begin{tabular}{lc}
\toprule
Hyperparameter & Value \\
\midrule
RL Algorithm & GiGPO \\
Learning Rate & $5 \times 10^{-6}$ \\
Train Batch Size & 256 \\
PPO Mini Batch Size & 256 \\
PPO Micro Batch Size & 32  \\
Group Size ($G$) & 8 \\
Max Response Length & 512 \\
KL Coefficient & 0.01 \\
Clip Range ($\epsilon$) & 0.2 \\
Steps & 150 \\
\bottomrule
\end{tabular}
\end{table}

\subsection{DeepSeek-Distilled-Qwen-7B (GRPO, Skywork-OR1)}
\label{app:deepseek_setup}

To test generalization beyond the Qwen model families, we train DeepSeek-R1-Distill-Qwen-7B (28 layers), a model distilled from DeepSeek-R1 into the Qwen architecture, using GRPO on the Skywork-OR1 mathematics dataset (approximately 48K problems). The reward is binary answer-matching. Table~\ref{tab:hp_deepseek} lists the hyperparameters.
Similar to the Qwen3 experiments, for the full-parameter baselines, we tune the learning rate over $\{1 \times 10^{-6},\, 3 \times 10^{-6},\, 5 \times 10^{-6},\, 1 \times 10^{-5}\}$ and report the best result (with learning rate $= 5 \times 10^{-6}$);  For single-layer training, we use a learning rate of $5 \times 10^{-6}$, which aligns with the full-parameter training.

\begin{table}[h]
\centering
\caption{Training hyperparameters for DeepSeek-Distilled-Qwen-7B (GRPO, Skywork-OR1).}
\label{tab:hp_deepseek}
\vspace{2mm}
\begin{tabular}{lc}
\toprule
Hyperparameter & Value \\
\midrule
RL Algorithm & GRPO \\
Learning Rate & $5 \times 10^{-6}$ \\
Train Batch Size & 256 \\
PPO Mini Batch Size & 128 \\
PPO Micro Batch Size & 2 \\
Group Size ($G$) & 8 \\
Max Response Length & 16384 \\
KL Coefficient & 0 \\
Clip Range ($\epsilon$) & 0.2 \\
Epochs & 8 \\
\bottomrule
\end{tabular}
\end{table}

\subsection{Training Datasets}
\label{app:datasets}

\paragraph{NuminaMath-CoT.} Our primary training dataset is derived from NuminaMath-CoT~\citep{numina_math_datasets}, a large-scale collection of approximately 860K competition-level math problems with chain-of-thought solutions, sourced from Chinese high school exams, US and international mathematics olympiads, and online mathematics forums. To improve training efficiency, we randomly downsample the dataset to 50K problems. We apply strict decontamination filtering to prevent test set leakage: for each of our evaluation benchmarks, we remove any training problem whose question text has a high n-gram overlap or semantic similarity with any test problem. The reward is binary: 1 if the model's final answer matches the ground truth, 0 otherwise.

\paragraph{DeepScaleR.} A curated mathematics dataset containing approximately 40K reasoning-intensive problems~\citep{deepscaler2025}, compiled from sources including AIME, AMC, and other competition archives. Used for cross-dataset validation in \S\ref{sec:consistency}. The same binary answer-matching reward is applied.

\paragraph{DeepCoder.} A coding dataset containing approximately 24K programming problems with test cases~\citep{deepcoder2025}, compiled from LiveCodeBench and Codeforces. Used for cross-task validation in \S\ref{sec:consistency}. The reward is based on execution correctness: 1 if the generated code passes all test cases, 0 otherwise.

\paragraph{Skywork-OR1.} A mathematics dataset containing approximately 48K problems. Used for training DeepSeek-Distilled-Qwen-7B in \S\ref{sec:generalization_family}. The same binary answer-matching reward is applied.

\paragraph{ALFWorld.} An environment-based agentic benchmark consisting of 2,435 household tasks across six categories. Each task requires the agent to interact with a simulated environment through text commands to achieve a goal (e.g., placing a heated object on a surface). The reward is binary: 1 if the task is completed successfully, 0 otherwise. Used for training the Qwen2.5-Instruct models in \S\ref{sec:generalization_family}.

\subsection{Learning Rate Ablation}
\label{app:lr_ablation}

A natural concern is whether the layer contribution rankings observed in \S\ref{sec:contribution} are artifacts of the learning rate choice---specifically, whether low-contribution layers might perform better with a larger learning rate, or whether high-contribution layers might lose their advantage under a different learning rate. We address this with the following ablation.

On Qwen3-1.7B-Base, we select the top-5 and bottom-5 layers by contribution ranking and retrain each of them individually with a $3\times$ higher learning rate ($1.5 \times 10^{-5}$ vs.\ the default $5 \times 10^{-6}$), keeping all other hyperparameters unchanged. Table~\ref{tab:lr_ablation} shows the results. The bottom-5 layers remain low-contribution under the boosted learning rate: their $\mathcal{C}$ values change by at most 0.02, and none of them approaches the high-contribution group. Similarly, the top-5 layers retain their high contribution under the same boosted learning rate. This confirms that the layer contribution profile is robust to the learning rate choice and is not an artifact of suboptimal hyperparameters, supporting the conclusion in \S\ref{sec:consistency} that layer contribution is an intrinsic property of the model.

\begin{table}[h]
\centering
\caption{Learning rate ablation on Qwen3-1.7B-Base. We compare layer contribution $\mathcal{C}$ of the top-5 and bottom-5 layers (by contribution ranking) under the default learning rate ($5 \times 10^{-6}$) and a $3\times$ boosted learning rate ($1.5 \times 10^{-5}$). Low-contribution layers remain low-contribution and high-contribution layers remain high-contribution under the higher learning rate.}
\label{tab:lr_ablation}
\vspace{2mm}
\small
\begin{tabular}{llcc}
\toprule
\textbf{Group} & \textbf{Layer} & $\mathcal{C}$ (LR=$5 \times 10^{-6}$) & $\mathcal{C}$ (LR=$1.5 \times 10^{-5}$) \\
\midrule
\multirow{5}{*}{Top-5}
& Layer 10 & 1.14 & 1.12 \\
& Layer 12 & 1.12 & 1.13 \\
& Layer 9  & 1.04 & 1.01 \\
& Layer 2  & 1.03 & 1.00 \\
& Layer 13 & 1.01 & 1.02 \\
\midrule
\multirow{5}{*}{Bottom-5}
& Layer 24 & 0.28 & 0.26 \\
& Layer 20 & 0.32 & 0.34 \\
& Layer 23 & 0.32 & 0.33 \\
& Layer 25 & 0.35 & 0.35 \\
& Layer 26 & 0.36 & 0.35 \\
\bottomrule
\end{tabular}
\end{table}

\section{Benchmark Selection Criteria}
\label{app:benchmarks}

We evaluate on 12 benchmarks grouped into four categories. Here we
describe each benchmark and our selection criteria.

\paragraph{Math (in-domain).}
\begin{itemize}[leftmargin=*]
    \item \textbf{MATH500}: 500 competition-level
    math problems spanning algebra, geometry, number theory, and more.
    \item \textbf{GSM8K}: 8.5K grade-school math
    word problems requiring multi-step arithmetic reasoning.
    \item \textbf{OlympiadBench}: Olympiad-level
    mathematics problems.
    \item \textbf{AMC}: Problems from the American
    Mathematics Competitions. \textbf{Considering the dataset is very small, we report the Average@32 results of evaluation on AMC.}
\end{itemize}

\paragraph{Code (out-of-distribution).}
\begin{itemize}[leftmargin=*]
    \item \textbf{HumanEval+}: Function-level
    code generation with augmented test cases.
    \item \textbf{MBPP}: Mostly Basic Python
    Programs, testing basic programming ability.
    \item \textbf{LiveCodeBench}: Recent
    competitive programming problems collected after model training
    cutoff dates.
\end{itemize}

\paragraph{Reasoning (out-of-distribution).}
\begin{itemize}[leftmargin=*]
    \item \textbf{GPQA-Diamond}: Graduate-level
    science questions curated by domain experts.
    \item \textbf{MMLU-Pro}: An enhanced version
    of MMLU with harder, more discriminative questions.
\end{itemize}

\paragraph{Language (out-of-distribution).}
\begin{itemize}[leftmargin=*]
    \item \textbf{C-Eval}: A comprehensive Chinese
    evaluation benchmark covering diverse subjects.
    \item \textbf{IFEval}: Instruction-following
    evaluation measuring the model's ability to follow specific
    formatting and content constraints.
    \item \textbf{MGSM}: Multilingual Grade School Math evaluates
    multilingual mathematical reasoning. 
\end{itemize}

\section{Full Per-Layer Results}
\label{app:full_results}

Tables~\ref{tab:full_1.7b},~\ref{tab:full_4b},
and~\ref{tab:full_8b} provide the complete per-layer evaluation
results of Qwen3 models for all 12 benchmarks on each model scale. Table ~\ref{tab:qwen25math_full} provide the complete per-layer evaluation
results for all 6 benchmarks on Qwen2.5-Math-1.5B.

\begin{table}[h]
\centering
\caption{Full per-layer results for Qwen3-1.7B-Base. $\mathcal{C}$ denotes layer contribution on math.}
\label{tab:full_1.7b}
\vspace{2mm}
\resizebox{\textwidth}{!}{%
\begin{tabular}{l|cccc|c|ccc|c|cc|c|ccc|c|c|c}
\toprule
\textbf{Setting} & \textbf{MATH500} & \textbf{GSM8K} & \textbf{Olymp.} & \textbf{AMC} & \textbf{Math} & \textbf{HE+} & \textbf{MBPP} & \textbf{LCB} & \textbf{Code} & \textbf{GPQA} & \textbf{MMLU-P} & \textbf{Reas.} & \textbf{C-Eval} & \textbf{IFEval} & \textbf{MGSM} & \textbf{Lang.} & \textbf{Overall} & $\mathcal{C}$ \\
\midrule
Base & 57.4 & 74.4 & 18.7 & 26.1 & 44.1 & 44.5 & 52.9 & 7.4 & 34.9 & 5.6 & 35.7 & 20.7 & 47.5 & 30.1 & 47.5 & 41.7 & 35.4 & 0.00 \\
Full & 64.0 & 82.0 & 26.9 & 30.2 & 50.8 & 43.3 & 46.3 & 10.9 & 33.5 & 5.0 & 40.1 & 22.6 & 56.2 & 32.4 & 56.0 & 48.2 & 38.8 & 1.00 \\
\midrule
Layer 0 & 63.4 & 81.7 & 23.6 & 31.9 & 50.1 & 46.3 & 55.6 & 10.9 & 37.6 & 5.0 & 39.5 & 22.3 & 54.9 & 31.6 & 51.8 & 46.1 & 39.0 & 0.89 \\
Layer 1 & 64.4 & 79.4 & 25.9 & 30.2 & 50.0 & 55.5 & 55.2 & 9.1 & 40.0 & 6.6 & 38.8 & 22.7 & 55.4 & 34.9 & 51.0 & 47.1 & 39.9 & 0.87 \\
Layer 2 & 67.8 & 80.0 & 26.1 & 30.5 & 51.1 & 42.7 & 53.7 & 10.9 & 35.7 & 6.1 & 39.5 & 22.8 & 55.7 & 32.4 & 52.0 & 46.7 & 39.1 & 1.03 \\
Layer 3 & 62.2 & 79.8 & 24.6 & 30.0 & 49.1 & 47.0 & 49.4 & 12.6 & 36.3 & 4.0 & 39.1 & 21.6 & 54.4 & 35.7 & 50.9 & 47.0 & 38.5 & 0.75 \\
Layer 4 & 61.2 & 81.3 & 25.5 & 29.2 & 49.3 & 44.5 & 51.8 & 10.3 & 35.5 & 5.0 & 38.7 & 21.9 & 54.0 & 33.1 & 51.8 & 46.3 & 38.3 & 0.77 \\
Layer 5 & 63.4 & 79.8 & 24.4 & 29.8 & 49.3 & 41.5 & 53.7 & 8.0 & 34.4 & 4.5 & 39.0 & 21.8 & 52.5 & 31.2 & 51.6 & 45.1 & 37.7 & 0.78 \\
Layer 6 & 63.8 & 80.4 & 25.6 & 30.0 & 50.0 & 48.8 & 51.0 & 9.1 & 36.3 & 3.5 & 39.0 & 21.3 & 54.4 & 31.6 & 53.5 & 46.5 & 38.5 & 0.87 \\
Layer 7 & 64.0 & 80.1 & 24.9 & 29.0 & 49.5 & 47.0 & 53.7 & 13.7 & 38.1 & 6.1 & 38.8 & 22.4 & 54.5 & 31.4 & 53.6 & 46.5 & 39.1 & 0.80 \\
Layer 8 & 62.4 & 79.8 & 25.0 & 28.7 & 49.0 & 45.7 & 51.8 & 10.3 & 35.9 & 3.5 & 39.0 & 21.3 & 51.3 & 29.9 & 53.5 & 44.9 & 37.8 & 0.72 \\
Layer 9 & 65.8 & 81.7 & 24.6 & 32.4 & 51.1 & 45.7 & 51.4 & 10.9 & 36.0 & 4.0 & 39.8 & 21.9 & 54.7 & 31.4 & 54.6 & 46.9 & 39.0 & 1.04 \\
Layer 10 & 68.6 & 80.5 & 27.3 & 30.8 & 51.8 & 40.9 & 53.7 & 9.1 & 34.6 & 5.0 & 38.7 & 21.9 & 55.6 & 31.1 & 54.9 & 47.2 & 38.9 & 1.14 \\
Layer 11 & 64.0 & 81.8 & 26.1 & 29.4 & 50.3 & 43.3 & 55.6 & 12.0 & 37.0 & 2.5 & 39.0 & 20.8 & 54.2 & 31.2 & 54.1 & 46.5 & 38.6 & 0.92 \\
Layer 12 & 65.6 & 81.3 & 27.3 & 32.4 & 51.6 & 42.7 & 52.9 & 13.1 & 36.2 & 7.1 & 39.4 & 23.2 & 55.7 & 32.2 & 55.2 & 47.7 & 39.7 & 1.12 \\
Layer 13 & 64.6 & 79.7 & 28.3 & 31.0 & 50.9 & 34.1 & 53.7 & 11.4 & 33.1 & 4.0 & 40.5 & 22.3 & 57.4 & 30.1 & 55.0 & 47.5 & 38.4 & 1.01 \\
Layer 14 & 64.4 & 79.6 & 27.1 & 30.0 & 50.3 & 40.9 & 53.7 & 10.3 & 34.9 & 4.0 & 40.4 & 22.2 & 55.4 & 29.6 & 53.6 & 46.2 & 38.4 & 0.92 \\
Layer 15 & 63.2 & 81.5 & 26.1 & 31.4 & 50.5 & 34.8 & 55.2 & 14.3 & 34.8 & 6.6 & 38.8 & 22.7 & 56.9 & 31.8 & 55.3 & 48.0 & 39.0 & 0.95 \\
Layer 16 & 64.8 & 80.9 & 23.0 & 30.8 & 49.9 & 37.2 & 52.5 & 8.6 & 32.8 & 3.0 & 39.4 & 21.2 & 56.0 & 32.4 & 54.1 & 47.5 & 37.8 & 0.85 \\
Layer 17 & 63.6 & 79.5 & 26.2 & 29.6 & 49.7 & 31.7 & 51.0 & 12.6 & 31.8 & 8.6 & 39.3 & 23.9 & 55.3 & 29.4 & 51.8 & 45.5 & 37.7 & 0.83 \\
Layer 18 & 61.8 & 78.5 & 24.3 & 29.7 & 48.6 & 34.1 & 49.4 & 12.6 & 32.0 & 6.1 & 39.3 & 22.7 & 52.4 & 29.8 & 51.9 & 44.7 & 37.0 & 0.66 \\
Layer 19 & 59.8 & 77.9 & 23.1 & 29.6 & 47.6 & 34.8 & 51.4 & 8.6 & 31.6 & 4.5 & 38.0 & 21.3 & 51.8 & 28.1 & 52.4 & 44.1 & 36.1 & 0.52 \\
Layer 20 & 60.6 & 74.9 & 21.6 & 28.1 & 46.3 & 31.7 & 51.0 & 9.7 & 30.8 & 6.1 & 37.2 & 21.6 & 52.4 & 29.6 & 50.0 & 44.0 & 35.7 & 0.32 \\
Layer 21 & 60.6 & 75.0 & 24.3 & 27.4 & 46.8 & 38.4 & 47.9 & 11.4 & 32.6 & 7.1 & 38.4 & 22.8 & 52.1 & 28.1 & 51.2 & 43.8 & 36.5 & 0.40 \\
Layer 22 & 60.6 & 76.5 & 24.3 & 27.9 & 47.3 & 35.4 & 51.8 & 8.6 & 31.9 & 4.5 & 36.7 & 20.6 & 51.9 & 28.8 & 51.9 & 44.2 & 36.0 & 0.47 \\
Layer 23 & 60.6 & 73.5 & 22.4 & 28.6 & 46.3 & 30.5 & 54.5 & 10.3 & 31.8 & 5.6 & 37.2 & 21.4 & 52.5 & 30.5 & 50.2 & 44.4 & 36.0 & 0.32 \\
Layer 24 & 60.6 & 74.8 & 21.2 & 27.6 & 46.1 & 29.9 & 50.6 & 11.4 & 30.6 & 6.1 & 37.1 & 21.6 & 52.8 & 29.8 & 50.0 & 44.2 & 35.6 & 0.28 \\
Layer 25 & 60.8 & 74.2 & 22.7 & 28.3 & 46.5 & 24.4 & 51.4 & 9.7 & 28.5 & 6.6 & 37.7 & 22.1 & 50.2 & 30.9 & 50.6 & 43.9 & 35.3 & 0.35 \\
Layer 26 & 60.6 & 74.4 & 23.4 & 27.8 & 46.5 & 29.3 & 51.0 & 9.1 & 29.8 & 4.0 & 36.8 & 20.4 & 52.4 & 27.0 & 50.2 & 43.2 & 35.0 & 0.36 \\
Layer 27 & 61.8 & 75.8 & 25.0 & 28.3 & 47.7 & 25.0 & 51.0 & 9.1 & 28.4 & 4.5 & 37.5 & 21.0 & 51.6 & 28.6 & 50.6 & 43.6 & 35.2 & 0.54 \\
\bottomrule
\end{tabular}
}
\end{table}

\begin{table}[h]
\centering
\caption{Full per-layer results for Qwen3-4B-Base. $\mathcal{C}$ denotes layer contribution on math.}
\label{tab:full_4b}
\vspace{2mm}
\resizebox{\textwidth}{!}{%
\begin{tabular}{l|cccc|c|ccc|c|cc|c|ccc|c|c|c}
\toprule
\textbf{Setting} & \textbf{MATH500} & \textbf{GSM8K} & \textbf{Olymp.} & \textbf{AMC} & \textbf{Math} & \textbf{HE+} & \textbf{MBPP} & \textbf{LCB} & \textbf{Code} & \textbf{GPQA} & \textbf{MMLU-P} & \textbf{Reas.} & \textbf{C-Eval} & \textbf{IFEval} & \textbf{MGSM} & \textbf{Lang.} & \textbf{Overall} & $\mathcal{C}$ \\
\midrule
Base & 65.2 & 75.4 & 27.6 & 40.5 & 52.2 & 68.3 & 44.8 & 11.4 & 41.5 & 5.1 & 52.5 & 28.8 & 69.9 & 39.7 & 63.2 & 57.6 & 45.0 & 0.00 \\
Full & 77.2 & 91.9 & 38.4 & 47.1 & 63.7 & 69.5 & 63.0 & 13.7 & 48.8 & 6.6 & 58.1 & 32.4 & 71.7 & 40.5 & 76.5 & 62.9 & 51.9 & 1.00 \\
\midrule
Layer 0 & 77.8 & 88.3 & 35.3 & 42.8 & 61.0 & 72.6 & 68.9 & 13.7 & 51.7 & 4.5 & 56.8 & 30.7 & 68.9 & 48.8 & 71.6 & 63.1 & 51.6 & 0.77 \\
Layer 1 & 74.0 & 87.7 & 36.4 & 42.7 & 60.2 & 70.7 & 67.3 & 12.6 & 50.2 & 6.1 & 55.2 & 30.6 & 71.1 & 46.2 & 72.9 & 63.4 & 51.1 & 0.70 \\
Layer 2 & 73.8 & 87.7 & 35.3 & 42.4 & 59.8 & 69.5 & 63.0 & 14.9 & 49.1 & 8.6 & 55.3 & 31.9 & 69.9 & 45.7 & 71.3 & 62.3 & 50.8 & 0.66 \\
Layer 3 & 74.4 & 87.5 & 36.3 & 42.5 & 60.2 & 70.1 & 57.2 & 15.4 & 47.6 & 9.1 & 55.1 & 32.1 & 68.6 & 44.4 & 72.1 & 61.7 & 50.4 & 0.70 \\
Layer 4 & 74.2 & 88.0 & 37.9 & 43.1 & 60.8 & 67.1 & 61.1 & 15.4 & 47.9 & 4.5 & 54.7 & 29.6 & 68.4 & 43.4 & 72.7 & 61.5 & 49.9 & 0.75 \\
Layer 5 & 76.6 & 86.3 & 37.0 & 44.3 & 61.1 & 65.2 & 61.1 & 13.1 & 46.5 & 8.6 & 55.5 & 32.0 & 69.3 & 46.4 & 71.5 & 62.4 & 50.5 & 0.77 \\
Layer 6 & 77.0 & 89.5 & 38.8 & 44.4 & 62.4 & 68.9 & 63.8 & 14.9 & 49.2 & 5.0 & 55.9 & 30.5 & 71.5 & 43.1 & 74.4 & 63.0 & 51.3 & 0.89 \\
Layer 7 & 73.8 & 89.4 & 39.1 & 44.1 & 61.6 & 68.9 & 64.2 & 14.9 & 49.3 & 7.6 & 55.8 & 31.7 & 72.0 & 41.6 & 73.3 & 62.3 & 51.2 & 0.82 \\
Layer 8 & 75.2 & 90.5 & 36.6 & 45.0 & 61.8 & 68.3 & 65.0 & 16.0 & 49.8 & 5.6 & 55.4 & 30.5 & 70.7 & 42.0 & 74.2 & 62.3 & 51.1 & 0.84 \\
Layer 9 & 77.2 & 89.3 & 37.0 & 44.9 & 62.1 & 68.9 & 61.5 & 13.7 & 48.0 & 4.0 & 55.9 & 30.0 & 70.8 & 42.9 & 73.8 & 62.5 & 50.7 & 0.87 \\
Layer 10 & 76.2 & 90.8 & 39.6 & 44.9 & 62.8 & 68.9 & 61.9 & 12.0 & 47.6 & 6.6 & 56.1 & 31.4 & 70.3 & 43.4 & 75.3 & 63.0 & 51.2 & 0.93 \\
Layer 11 & 76.6 & 90.5 & 36.2 & 47.6 & 62.7 & 73.2 & 66.5 & 14.3 & 51.3 & 11.1 & 56.4 & 33.8 & 71.0 & 41.6 & 75.8 & 62.8 & 52.7 & 0.92 \\
Layer 12 & 80.8 & 90.4 & 37.0 & 45.6 & 63.4 & 69.5 & 66.5 & 13.7 & 49.9 & 7.6 & 56.8 & 32.2 & 70.2 & 42.9 & 75.3 & 62.8 & 52.1 & 0.98 \\
Layer 13 & 76.8 & 89.5 & 38.1 & 44.8 & 62.3 & 69.5 & 66.5 & 13.1 & 49.7 & 6.6 & 56.8 & 31.7 & 70.4 & 41.6 & 76.1 & 62.7 & 51.6 & 0.88 \\
Layer 14 & 78.4 & 90.3 & 39.9 & 46.5 & 63.8 & 73.8 & 70.0 & 14.9 & 52.9 & 5.0 & 58.1 & 31.6 & 71.8 & 42.1 & 76.0 & 63.3 & 52.9 & 1.02 \\
Layer 15 & 78.0 & 90.8 & 39.0 & 46.2 & 63.5 & 70.1 & 64.6 & 13.7 & 49.5 & 6.1 & 57.7 & 31.9 & 72.1 & 46.0 & 76.6 & 64.9 & 52.4 & 0.98 \\
Layer 16 & 79.4 & 92.0 & 40.3 & 45.5 & 64.3 & 75.0 & 66.5 & 14.3 & 51.9 & 8.1 & 57.9 & 33.0 & 73.8 & 42.3 & 77.1 & 64.4 & 53.4 & 1.06 \\
Layer 17 & 78.4 & 91.0 & 38.8 & 45.9 & 63.5 & 74.4 & 65.0 & 17.1 & 52.2 & 6.1 & 57.9 & 32.0 & 71.5 & 43.4 & 77.1 & 64.0 & 52.9 & 0.99 \\
Layer 18 & 78.6 & 90.5 & 37.9 & 45.5 & 63.1 & 67.1 & 69.7 & 15.4 & 50.7 & 7.1 & 57.5 & 32.3 & 72.6 & 43.6 & 77.0 & 64.4 & 52.6 & 0.96 \\
Layer 19 & 78.0 & 91.4 & 40.3 & 45.6 & 63.8 & 70.1 & 66.2 & 14.3 & 50.2 & 6.1 & 57.7 & 31.9 & 73.1 & 42.0 & 76.6 & 63.9 & 52.5 & 1.02 \\
Layer 20 & 79.0 & 90.7 & 39.0 & 43.6 & 63.1 & 68.9 & 66.2 & 16.6 & 50.5 & 6.6 & 58.1 & 32.3 & 71.3 & 41.6 & 75.8 & 62.9 & 52.2 & 0.95 \\
Layer 21 & 76.2 & 89.9 & 39.6 & 45.5 & 62.8 & 65.2 & 67.3 & 17.7 & 50.1 & 7.6 & 57.5 & 32.6 & 74.2 & 41.6 & 75.6 & 63.8 & 52.3 & 0.93 \\
Layer 22 & 78.0 & 90.8 & 39.3 & 46.5 & 63.6 & 68.3 & 64.2 & 16.0 & 49.5 & 6.1 & 58.0 & 32.0 & 70.7 & 41.2 & 75.9 & 62.6 & 51.9 & 1.00 \\
Layer 23 & 76.8 & 90.5 & 37.5 & 44.9 & 62.4 & 67.1 & 65.4 & 16.0 & 49.5 & 7.6 & 56.8 & 32.2 & 70.7 & 40.7 & 75.5 & 62.3 & 51.6 & 0.89 \\
Layer 24 & 77.0 & 89.5 & 38.5 & 43.6 & 62.2 & 72.0 & 58.4 & 12.0 & 47.4 & 7.6 & 56.3 & 31.9 & 70.7 & 39.6 & 74.8 & 61.7 & 50.8 & 0.87 \\
Layer 25 & 76.6 & 89.0 & 36.0 & 44.4 & 61.5 & 70.7 & 57.2 & 16.0 & 48.0 & 7.6 & 56.4 & 32.0 & 71.5 & 40.9 & 73.9 & 62.1 & 50.9 & 0.81 \\
Layer 26 & 73.8 & 89.7 & 35.9 & 44.3 & 60.9 & 69.5 & 56.0 & 16.6 & 47.4 & 5.0 & 56.4 & 30.7 & 71.9 & 40.3 & 75.0 & 62.4 & 50.3 & 0.76 \\
Layer 27 & 75.6 & 89.5 & 34.2 & 43.5 & 60.7 & 70.7 & 55.6 & 13.1 & 46.5 & 7.1 & 56.7 & 31.9 & 70.2 & 39.7 & 73.7 & 61.2 & 50.1 & 0.74 \\
Layer 28 & 75.6 & 89.0 & 35.7 & 43.2 & 60.9 & 68.3 & 65.8 & 14.3 & 49.4 & 3.0 & 57.0 & 30.0 & 70.4 & 37.9 & 75.0 & 61.1 & 50.4 & 0.76 \\
Layer 29 & 75.4 & 87.3 & 36.0 & 43.8 & 60.6 & 68.9 & 45.1 & 14.9 & 43.0 & 7.1 & 56.0 & 31.5 & 69.5 & 33.5 & 74.3 & 59.1 & 48.6 & 0.74 \\
Layer 30 & 72.8 & 89.1 & 36.7 & 43.6 & 60.6 & 68.9 & 44.0 & 14.9 & 42.6 & 7.1 & 57.0 & 32.1 & 68.6 & 32.9 & 73.7 & 58.4 & 48.4 & 0.73 \\
Layer 31 & 76.6 & 88.0 & 39.4 & 45.6 & 62.4 & 72.0 & 48.2 & 13.1 & 44.4 & 8.1 & 57.1 & 32.6 & 66.6 & 31.1 & 74.2 & 57.3 & 49.2 & 0.89 \\
Layer 32 & 75.4 & 88.0 & 37.2 & 41.9 & 60.6 & 63.4 & 29.2 & 16.0 & 36.2 & 6.6 & 56.1 & 31.3 & 66.8 & 31.8 & 73.9 & 57.5 & 46.4 & 0.74 \\
Layer 33 & 75.2 & 87.7 & 37.2 & 42.8 & 60.7 & 67.7 & 49.8 & 14.3 & 43.9 & 7.1 & 56.5 & 31.8 & 68.5 & 32.0 & 74.7 & 58.4 & 48.7 & 0.75 \\
Layer 34 & 76.4 & 88.3 & 35.1 & 43.1 & 60.7 & 61.6 & 46.3 & 13.7 & 40.5 & 4.5 & 55.2 & 29.9 & 69.8 & 30.5 & 73.4 & 57.9 & 47.3 & 0.75 \\
Layer 35 & 76.4 & 89.6 & 38.5 & 45.2 & 62.4 & 61.0 & 47.1 & 13.7 & 40.6 & 3.0 & 55.2 & 29.1 & 70.1 & 29.4 & 73.6 & 57.7 & 47.5 & 0.89 \\
\bottomrule
\end{tabular}
}
\end{table}

\begin{table}[h]
\centering
\caption{Full per-layer results for Qwen3-8B-Base. $\mathcal{C}$ denotes layer contribution on math.}
\label{tab:full_8b}
\vspace{2mm}
\resizebox{\textwidth}{!}{%
\begin{tabular}{l|cccc|c|ccc|c|cc|c|ccc|c|c|c}
\toprule
\textbf{Setting} & \textbf{MATH500} & \textbf{GSM8K} & \textbf{Olymp.} & \textbf{AMC} & \textbf{Math} & \textbf{HE+} & \textbf{MBPP} & \textbf{LCB} & \textbf{Code} & \textbf{GPQA} & \textbf{MMLU-P} & \textbf{Reas.} & \textbf{C-Eval} & \textbf{IFEval} & \textbf{MGSM} & \textbf{Lang.} & \textbf{Overall} & $\mathcal{C}$ \\
\midrule
Base & 71.8 & 82.0 & 36.6 & 41.7 & 58.0 & 67.1 & 66.9 & 17.1 & 50.4 & 6.6 & 57.7 & 32.2 & 71.5 & 46.2 & 54.8 & 57.5 & 49.5 & 0.00 \\
Full & 80.0 & 92.3 & 42.8 & 50.8 & 66.5 & 75.6 & 67.3 & 18.3 & 53.7 & 8.1 & 63.0 & 35.5 & 70.1 & 47.7 & 73.3 & 63.7 & 54.9 & 1.00 \\
\midrule
Layer 0 & 61.8 & 79.6 & 31.0 & 42.4 & 53.7 & 66.5 & 49.0 & 16.6 & 44.0 & 7.6 & 60.3 & 34.0 & 73.1 & 46.4 & 69.5 & 63.0 & 48.7 & -0.51 \\
Layer 1 & 72.2 & 84.7 & 36.3 & 45.8 & 59.7 & 72.6 & 65.8 & 16.0 & 51.4 & 8.6 & 60.1 & 34.3 & 72.3 & 49.4 & 67.3 & 63.0 & 52.1 & 0.20 \\
Layer 2 & 72.8 & 84.6 & 38.5 & 43.0 & 59.7 & 73.2 & 67.3 & 13.7 & 51.4 & 7.1 & 60.3 & 33.7 & 71.9 & 45.5 & 60.8 & 59.4 & 51.1 & 0.20 \\
Layer 3 & 77.8 & 89.3 & 37.0 & 48.6 & 63.2 & 78.0 & 68.9 & 18.3 & 55.1 & 9.1 & 60.0 & 34.6 & 72.7 & 47.3 & 68.1 & 62.7 & 53.9 & 0.61 \\
Layer 4 & 73.6 & 84.8 & 39.0 & 49.6 & 61.7 & 76.8 & 70.8 & 18.3 & 55.3 & 9.6 & 60.4 & 35.0 & 76.2 & 48.4 & 62.9 & 62.5 & 53.6 & 0.44 \\
Layer 5 & 76.8 & 88.7 & 40.4 & 49.0 & 63.7 & 78.0 & 68.5 & 14.9 & 53.8 & 10.1 & 60.8 & 35.5 & 75.4 & 48.6 & 65.3 & 63.1 & 54.0 & 0.67 \\
Layer 6 & 79.0 & 90.1 & 40.0 & 50.7 & 65.0 & 76.8 & 73.9 & 17.7 & 56.2 & 8.1 & 62.3 & 35.2 & 74.4 & 54.5 & 68.5 & 65.8 & 55.5 & 0.82 \\
Layer 7 & 79.4 & 88.5 & 40.1 & 48.6 & 64.2 & 73.8 & 73.5 & 16.6 & 54.6 & 9.6 & 61.7 & 35.6 & 73.7 & 49.5 & 66.7 & 63.3 & 54.4 & 0.72 \\
Layer 8 & 77.0 & 89.0 & 41.5 & 50.9 & 64.6 & 72.0 & 71.6 & 19.4 & 54.3 & 7.1 & 61.8 & 34.4 & 73.8 & 46.6 & 65.6 & 62.0 & 53.8 & 0.78 \\
Layer 9 & 80.4 & 90.9 & 41.2 & 51.5 & 66.0 & 71.3 & 72.4 & 17.1 & 53.6 & 5.6 & 61.2 & 33.4 & 71.9 & 46.6 & 65.4 & 61.3 & 53.6 & 0.94 \\
Layer 10 & 78.6 & 90.6 & 41.0 & 51.5 & 65.4 & 79.3 & 69.3 & 16.6 & 55.0 & 9.1 & 61.8 & 35.4 & 73.7 & 49.4 & 72.2 & 65.1 & 55.3 & 0.87 \\
Layer 11 & 75.4 & 84.5 & 37.2 & 49.9 & 61.7 & 79.9 & 71.2 & 17.1 & 56.1 & 9.1 & 61.7 & 35.4 & 71.0 & 47.7 & 52.6 & 57.1 & 52.6 & 0.44 \\
Layer 12 & 77.8 & 90.8 & 39.4 & 52.2 & 65.1 & 76.8 & 69.7 & 18.3 & 54.9 & 5.6 & 61.6 & 33.6 & 75.8 & 48.1 & 70.2 & 64.7 & 54.6 & 0.83 \\
Layer 13 & 80.8 & 90.5 & 41.0 & 51.7 & 66.0 & 78.0 & 68.5 & 15.4 & 54.0 & 8.1 & 62.0 & 35.0 & 74.6 & 48.1 & 65.4 & 62.7 & 54.4 & 0.94 \\
Layer 14 & 81.6 & 92.6 & 41.2 & 51.8 & 66.8 & 80.5 & 70.0 & 16.6 & 55.7 & 5.0 & 62.4 & 33.7 & 75.2 & 51.0 & 76.3 & 67.5 & 55.9 & 1.03 \\
Layer 15 & 79.8 & 92.8 & 40.6 & 52.7 & 66.5 & 78.7 & 73.5 & 18.3 & 56.8 & 5.0 & 63.0 & 34.0 & 77.0 & 51.9 & 77.8 & 68.9 & 56.5 & 1.00 \\
Layer 16 & 80.4 & 91.8 & 44.1 & 52.0 & 67.1 & 76.2 & 72.4 & 14.9 & 54.5 & 7.6 & 63.5 & 35.5 & 76.1 & 52.5 & 77.8 & 68.8 & 56.5 & 1.07 \\
Layer 17 & 81.6 & 92.5 & 41.6 & 52.5 & 67.1 & 76.2 & 70.4 & 16.0 & 54.2 & 6.6 & 63.1 & 34.9 & 74.9 & 49.5 & 76.6 & 67.0 & 55.8 & 1.07 \\
Layer 18 & 80.6 & 92.0 & 39.6 & 51.2 & 65.8 & 76.8 & 71.2 & 13.7 & 53.9 & 9.6 & 63.1 & 36.3 & 75.0 & 45.7 & 74.0 & 64.9 & 55.2 & 0.92 \\
Layer 19 & 79.6 & 91.8 & 43.0 & 50.7 & 66.3 & 72.6 & 68.9 & 18.3 & 53.2 & 9.6 & 62.6 & 36.1 & 72.8 & 48.2 & 72.8 & 64.6 & 55.1 & 0.97 \\
Layer 20 & 77.2 & 92.3 & 41.2 & 51.3 & 65.5 & 68.9 & 64.6 & 17.7 & 50.4 & 9.1 & 62.2 & 35.7 & 73.3 & 44.9 & 70.8 & 63.0 & 53.6 & 0.88 \\
Layer 21 & 79.8 & 92.4 & 41.5 & 49.9 & 65.9 & 71.3 & 68.1 & 15.4 & 51.6 & 8.6 & 62.0 & 35.3 & 74.3 & 47.5 & 69.0 & 63.6 & 54.1 & 0.93 \\
Layer 22 & 80.8 & 91.2 & 41.9 & 51.2 & 66.3 & 70.7 & 64.6 & 17.1 & 50.8 & 6.1 & 62.3 & 34.2 & 74.0 & 45.3 & 72.7 & 64.0 & 53.8 & 0.98 \\
Layer 23 & 80.8 & 90.2 & 42.5 & 50.4 & 66.0 & 70.1 & 62.6 & 17.1 & 50.0 & 9.1 & 61.4 & 35.2 & 69.0 & 42.1 & 68.6 & 59.9 & 52.8 & 0.94 \\
Layer 24 & 79.8 & 91.1 & 41.2 & 49.7 & 65.5 & 70.1 & 60.7 & 16.6 & 49.1 & 7.1 & 60.9 & 34.0 & 70.3 & 43.6 & 66.7 & 60.2 & 52.2 & 0.88 \\
Layer 25 & 78.0 & 89.4 & 39.4 & 48.0 & 63.7 & 70.7 & 58.8 & 18.9 & 49.4 & 7.6 & 60.8 & 34.2 & 72.4 & 44.0 & 69.0 & 61.8 & 52.3 & 0.67 \\
Layer 26 & 77.0 & 89.2 & 38.2 & 49.0 & 63.3 & 68.9 & 58.8 & 18.9 & 48.8 & 7.1 & 60.6 & 33.8 & 69.2 & 43.4 & 65.0 & 59.2 & 51.3 & 0.63 \\
Layer 27 & 80.8 & 88.8 & 41.0 & 49.2 & 65.0 & 74.4 & 59.1 & 16.0 & 49.8 & 8.6 & 61.0 & 34.8 & 69.9 & 44.5 & 66.8 & 60.4 & 52.5 & 0.82 \\
Layer 28 & 79.0 & 88.6 & 41.2 & 48.3 & 64.3 & 72.6 & 58.8 & 17.1 & 49.5 & 7.1 & 60.9 & 34.0 & 70.6 & 43.4 & 67.8 & 60.6 & 52.1 & 0.74 \\
Layer 29 & 78.8 & 87.9 & 39.4 & 47.5 & 63.4 & 68.9 & 58.0 & 15.4 & 47.4 & 9.1 & 61.0 & 35.0 & 72.1 & 45.1 & 65.2 & 60.8 & 51.7 & 0.64 \\
Layer 30 & 78.6 & 88.2 & 38.1 & 48.8 & 63.4 & 68.9 & 56.4 & 13.7 & 46.3 & 9.6 & 60.9 & 35.3 & 69.8 & 46.0 & 63.3 & 59.7 & 51.2 & 0.64 \\
Layer 31 & 77.4 & 88.9 & 40.0 & 50.3 & 64.2 & 75.0 & 53.3 & 18.3 & 48.9 & 9.6 & 60.3 & 34.9 & 69.7 & 45.3 & 60.5 & 58.5 & 51.6 & 0.72 \\
Layer 32 & 71.8 & 85.4 & 38.1 & 46.1 & 60.3 & 54.9 & 47.1 & 15.4 & 39.1 & 9.1 & 58.0 & 33.5 & 66.5 & 46.0 & 61.5 & 58.0 & 47.8 & 0.27 \\
Layer 33 & 78.0 & 87.6 & 39.3 & 48.2 & 63.3 & 70.1 & 35.0 & 16.0 & 40.4 & 7.1 & 60.6 & 33.8 & 68.0 & 46.6 & 66.6 & 60.4 & 49.5 & 0.62 \\
Layer 34 & 77.6 & 87.0 & 39.7 & 48.8 & 63.3 & 60.4 & 40.5 & 17.1 & 39.3 & 7.1 & 59.6 & 33.3 & 63.1 & 38.3 & 73.5 & 58.3 & 48.6 & 0.62 \\
Layer 35 & 78.4 & 88.9 & 41.2 & 49.3 & 64.5 & 60.4 & 29.2 & 18.3 & 35.9 & 7.6 & 61.0 & 34.3 & 67.8 & 33.6 & 70.5 & 57.3 & 48.0 & 0.76 \\
\bottomrule
\end{tabular}
}
\end{table}

\begin{table}[h]
\centering
\caption{Full per-layer results for Qwen2.5-Math-1.5B (Dr.\,GRPO). $\mathcal{C}$ denotes
layer contribution on Avg.}
\label{tab:qwen25math_full}
\begin{tabular}{lcccccccc}
\toprule
Setting & AIME & AIME25 & AMC & MATH500 & Minerva & Olymp. & Avg & $\mathcal{C}$ \\
\midrule
Base    & 20.0 & 6.7  & 32.5 & 33.0 & 12.5 & 22.8 & 21.2 & 0.00 \\
Full    & 16.7 & 10.0 & 51.8 & 74.4 & 25.0 & 38.8 & 36.1 & 1.00 \\
\midrule
Layer 0  & 10.0 & 10.0 & 47.0 & 70.4 & 23.2 & 33.9 & 32.4 & 0.75 \\
Layer 1  & 10.0 & 6.7  & 49.4 & 68.2 & 20.2 & 31.6 & 31.0 & 0.66 \\
Layer 2  & 10.0 & 3.3  & 44.6 & 71.2 & 21.7 & 32.6 & 30.6 & 0.63 \\
Layer 3  & 20.0 & 10.0 & 45.8 & 69.8 & 21.3 & 33.0 & 33.3 & 0.81 \\
Layer 4  & 13.3 & 3.3  & 44.6 & 70.4 & 23.2 & 32.7 & 31.2 & 0.67 \\
Layer 5  & 16.7 & 3.3  & 44.6 & 69.4 & 22.4 & 32.1 & 31.4 & 0.68 \\
Layer 6  & 6.7  & 6.7  & 43.4 & 69.2 & 22.1 & 31.6 & 29.9 & 0.59 \\
Layer 7  & 6.7  & 6.7  & 42.2 & 69.0 & 20.6 & 30.2 & 29.2 & 0.54 \\
Layer 8  & 13.3 & 3.3  & 43.4 & 69.4 & 20.6 & 30.7 & 30.1 & 0.60 \\
Layer 9  & 20.0 & 3.3  & 43.4 & 69.8 & 23.2 & 33.0 & 32.1 & 0.73 \\
Layer 10 & 13.3 & 6.7  & 44.6 & 70.4 & 23.2 & 32.6 & 31.8 & 0.71 \\
Layer 11 & 16.7 & 3.3  & 44.6 & 69.6 & 21.7 & 32.3 & 31.4 & 0.68 \\
Layer 12 & 20.0 & 10.0 & 45.8 & 73.8 & 25.0 & 34.8 & 34.9 & 0.92 \\
Layer 13 & 13.3 & 10.0 & 45.8 & 69.6 & 22.8 & 34.4 & 32.6 & 0.77 \\
Layer 14 & 20.0 & 10.0 & 52.3 & 74.8 & 25.6 & 35.3 & \textbf{36.3} & \textbf{1.01} \\
Layer 15 & 16.7 & 10.0 & 43.4 & 73.8 & 26.1 & 36.7 & 34.4 & 0.89 \\
Layer 16 & 20.0 & 10.0 & 51.8 & 75.2 & 24.9 & 34.9 & 36.1 & 1.00 \\
Layer 17 & 16.7 & 6.7  & 49.4 & 73.4 & 24.6 & 35.3 & 34.3 & 0.88 \\
Layer 18 & 10.0 & 0.0  & 49.4 & 71.0 & 24.6 & 34.1 & 31.5 & 0.69 \\
Layer 19 & 10.0 & 0.0  & 45.8 & 70.4 & 22.8 & 31.7 & 30.1 & 0.60 \\
Layer 20 & 10.0 & 0.0  & 45.8 & 64.6 & 20.2 & 30.7 & 28.5 & 0.49 \\
Layer 21 & 13.3 & 3.3  & 45.8 & 67.2 & 19.1 & 29.6 & 29.7 & 0.57 \\
Layer 22 & 13.3 & 0.0  & 37.3 & 65.2 & 19.9 & 30.5 & 27.7 & 0.43 \\
Layer 23 & 10.0 & 3.3  & 38.6 & 64.0 & 19.9 & 29.3 & 27.5 & 0.42 \\
Layer 24 & 16.7 & 3.3  & 37.3 & 65.2 & 21.0 & 29.9 & 28.9 & 0.51 \\
Layer 25 & 16.7 & 3.3  & 41.0 & 62.4 & 19.1 & 28.1 & 28.4 & 0.48 \\
Layer 26 & 13.3 & 3.3  & 42.2 & 66.2 & 20.6 & 29.9 & 29.2 & 0.54 \\
Layer 27 & 10.0 & 6.7  & 42.2 & 69.4 & 24.3 & 32.3 & 30.8 & 0.64 \\
\bottomrule
\end{tabular}
\end{table}

\newpage

\end{document}